\documentclass[journal=jcisd8,manuscript=article]{achemso}

\usepackage[version=3]{mhchem} 
\usepackage{subcaption}
\usepackage{xcolor}  
\usepackage{bm}
\usepackage{hyperref}
\usepackage{amssymb}
\usepackage[symbol]{footmisc}

\author{Marco~Hoffmann}
\author{Hans~Hasse}
\author{Fabian~Jirasek}\email{fabian.jirasek@rptu.de}
\affiliation{Laboratory of Engineering Thermodynamics, RPTU Kaiserslautern, Erwin-Schrödinger-Str. 44, 67663 Kaiserslautern, Germany}

\title{GRAPPA - A Hybrid Graph Neural Network for Predicting Pure Component Vapor Pressures}

\newcommand{\ps}{$p^\mathrm{s}$ }

\begin{document}

\begin{abstract}
Although the pure component vapor pressure is one of the most important properties for designing chemical processes, no broadly applicable, sufficiently accurate, and open-source prediction method has been available. To overcome this, we have developed GRAPPA - a hybrid graph neural network for predicting vapor pressures of pure components. GRAPPA enables the prediction of the vapor pressure curve of basically any organic molecule, requiring only the molecular structure as input. The new model consists of three parts: A graph attention network for the message passing step, a pooling function that captures long-range interactions, and a prediction head that yields the component-specific parameters of the Antoine equation, from which the vapor pressure can readily and consistently be calculated for any temperature. We have trained and evaluated GRAPPA on experimental vapor pressure data of almost 25,000 pure components. We found excellent prediction accuracy for unseen components, outperforming state-of-the-art group contribution methods and other machine learning approaches in applicability and accuracy. The trained model and its code are fully disclosed, and GRAPPA is directly applicable via the interactive website \texttt{ml-prop.mv.rptu.de}.
\end{abstract}

\section{Introduction}
The vapor pressure \ps of pure components is a key property in designing and optimizing many chemical processes. However, since experimental data for the vapor pressure are, by far, not available for all relevant components and at all relevant temperatures, prediction methods for this property are paramount. We can thereby distinguish between two types of prediction problems:
\begin{enumerate}
    \item The interpolation or extrapolation of the vapor pressure to unstudied temperatures for components for which some experimental data (at other temperatures) are available.
    \item The prediction of the vapor pressure of components for which no data (at any temperature) are available.
\end{enumerate}

The first, much easier, type of prediction problem is usually solved by employing semi-empirical correlations of the vapor pressure as a function of the temperature, of which the Antoine equation is a simple example:
\begin{equation}
\label{eq:antoine_equation}
    \ln \left( p^\mathrm{s} / \mathrm{kPa} \right) = A - \frac{B}{C + T / \mathrm{K}}
\end{equation}
The Antoine equation is a good compromise between simplicity and accuracy and therefore widely used - and there are databases containing the Antoine parameters for several thousand pure components in the literature~\cite{DDB2024, Yaws2015}. However, for the largest share of the pure components from the chemical space, no experimental data for the vapor pressure are available, so fitting semi-empirical correlations like Eq.~(\ref{eq:antoine_equation}) is infeasible. In these cases, prediction methods that can generalize over components are needed. In the following, we give a brief overview of such methods from the literature. First, we discuss established categories, which we divide into corresponding states, group contribution, quantitative structure-property relations (QSPR), and equation of state (EoS) based methods. Then, we will discuss recent machine learning (ML) developments. We acknowledge some ambiguities and that some methods might fit into multiple categories.

\textbf{Corresponding states methods.}\cite{Lee1975, Ambrose1989, Riedel1954}
These methods follow the idea that the thermophysical properties of many components are similar if the properties are reduced using the components' critical properties. Consequently, corresponding states methods are based on the availability of critical data for the components of interest, specifically the critical pressure $p_\mathrm{c}$ and the critical temperature $T_\mathrm{c}$. Some methods also require the acentric factor $\omega$ (representing one additional data point of the vapor pressure curve). Furthermore, these methods are often limited to components of similar chemical structure. For example, the method of Ambrose and Walton \cite{Ambrose1989} only applies to alkanes and 1-alcohols. For these reasons, corresponding states methods for predicting vapor pressures usually have a relatively small scope, hampering their practical applicability.

\textbf{Group contribution methods.} \cite{Macknick1979, Edwards1981, Burkhard1985, Tu1994, Sawaya2004, Asher2006, Pankow2008, Nannoolal2008, Moller2008, Ceriani2013, Rezakazemi2013, Wang2015a}
The idea behind group contribution methods (GCMs) is to decompose a component into pre-defined structural groups (molecular building blocks) and to model the properties of components as a function of their group composition. Usually, additivity of the contributions of the individual structural groups to the value of the target property of the component is assumed (occasionally, also group interactions are considered), and group-specific parameters are fitted to the training data. The target property can then, in principle, be derived for any component that can be built from the parameterized structural groups. Theoretically, this allows GCMs to predict vapor pressures only from molecular structures. However, in practice, most methods require one additional experimental data point, such as, e.g., the normal boiling temperature \cite{Nannoolal2008, Moller2008, Sawaya2004} or the critical temperature \cite{Wang2015a}. Only very few GCMs, e.g., those by Pankow and Asher \cite{Pankow2008} and Tu \cite{Tu1994}, apply to a wide range of components since they do not require any experimental data. Further limitations arise from the chemical space covered by the structural groups considered in the GCMs.

\textbf{QSPR and early machine learning methods.}\cite{Beck2000, Gharagheizi2012, Katritzky2007}
The idea of quantitative structure-property relations (QSPR) methods \cite{Katritzky1995} is to identify the molecular descriptors that are most informative for the thermophysical property to be predicted and to find a suitable correlation between these descriptors and the target property. These molecular descriptors are associated with the chemical structure of the molecule. Some simple examples are the dipole moment, the molecular volume and information on the connectivity of the atoms in the molecule\cite{Katritzky2007}. The QSPR approach has been used to develop methods for predicting vapor pressures in different ways, e.g., the descriptors have been used as input for a multilinear regression \cite{Katritzky2007} or the training of neural networks (NNs)\cite{Gharagheizi2012, Beck2000}.

\textbf{Equation of state based methods.} \cite{Hsieh2008, Wang2015, Liang2019, Habicht2023, Felton2024, Winter2023}
These methods derive the parameters of an EoS from the molecular structure (or certain descriptors of which) and use this EoS to calculate thermophysical properties, enabling the EoS to generalize over components. For example, Hsieh et al.~\cite{Hsieh2008} introduced a Peng-Robinson (PR) EoS parameterized using descriptors obtained from COSMO \cite{Klamt2000, Bell2020} calculations. With the resulting PR+COSMOSAC EoS \cite{Hsieh2008, Wang2015, Liang2019}, pure component vapor pressures can be predicted. There are also some recent works \cite{Habicht2023, Felton2024, Winter2023} that leverage EoS for pure component vapor pressure prediction, which are, however, discussed in the next section, as the focus of these works lies on the implementation of modern ML methods. Generally, in the EoS-based methods, the vapor pressure needs to be calculated iteratively, which is in contrast to the other methods discussed above that allow for a direct calculation. This makes them less practicable for an application in process simulators.

\textbf{Recent ML-based methods.} \cite{Habicht2023, Felton2024, Winter2023, Santana2024, Lin2024} Recently, many methods based on machine learning have been published to predict the thermophysical properties of pure components and mixtures \cite{Habicht2023, Felton2024, Winter2023, Santana2024, Lin2024, SanchezMedina2022, SanchezMedina2023, Ahmad2023, Qu2022, Aouichaoui2023, Aouichaoui2023a, Aouichaoui2023b, Hayer2024a, Hayer2025, Specht2024, Damay2021, Rittig2023, Rittig2023a}. Among them, several methods focus on predicting pure component vapor pressures, which all incorporate thermodynamic knowledge in their architecture and are thus not purely data-driven but hybrid methods\cite{Jirasek2023}. These methods can, e.g., be distinguished by the type of molecular embedding used as input for the prediction.

One option is to use extended connectivity fingerprints (ECFP) \cite{Rogers2010} as molecular embedding. For example, Habicht et al. \cite{Habicht2023} and Felton et al. \cite{Felton2024} trained NNs to regressed PC(P)-SAFT EoS\cite{Gross2001, Gross2005} parameters using ECFPs as input. The vapor pressures can then be calculated from the EoS. Another possibility is to derive the molecular embedding directly from the SMILES \cite{Weininger1988} string of a molecule, using, e.g., a transformer-based language model. This approach was followed by Winter et al. \cite{Winter2023}, who also predicted PC-SAFT parameters with their model; they did not only train on vapor pressure data but also on density data. 

Another approach is to derive a molecular embedding using a graph neural network (GNN). These have gained increasing interest for molecular property prediction in the last years, based, among others, on the initial work by Duvenaud et al.\cite{Duvenaud2015}. The underlying idea is that a molecule can intuitively be represented as a graph, with atoms as nodes and bonds as edges. Using a GNN, the node representations are iteratively updated \footnote[2]{In some architectures, the edge embeddings are updated instead, see Ref.\cite{Yang2019} for an example.} (using the so-called message passing) based on the surrounding nodes (and edges). Thus, each node is enriched with information on its chemical environment. The final graph embedding is then used to predict a target property. Because the parameters of the GNN are adjusted during the training process, the obtained embedding is explicitly tailored to the target property. This learnable embedding from GNNs significantly improves the "static" molecular fingerprints (such as ECFP). Given that GNNs take into account the chemical environment of the atoms in the molecule, they also have the potential to outperform classic GCMs, as the latter usually do not use the complete connectivity information of the structural groups in a component. For this reason, GNNs have already been adopted for the prediction of activity coefficients \cite{SanchezMedina2022, SanchezMedina2023, Rittig2023a}, solubility \cite{Ahmad2023}, normal boiling points \cite{Qu2022, Wang2023}, critical temperatures \cite{Wang2023} and several other thermophysical, safety-related, and environmental properties\cite{Aouichaoui2023b}. 

Very recently, two works were published in which GNNs were used to predict vapor pressures \cite{Santana2024, Lin2024}. These developments were carried out in parallel and independently from those on which we report here. The PUFFIN \cite{Santana2024} framework is built on a graph convolutional network (GCN) that was pre-trained to predict normal boiling points. In a transfer-learning approach, the embeddings of this pre-trained GCN were then used to predict Antoine parameters, from which the vapor pressures can be calculated. For training, Santana et al. \cite{Santana2024} used experimental data (at $298\,\mathrm{K}$) and synthetic data (at four other temperatures) of 1851 pure components. Their results prove that their model can extrapolate to unseen temperatures. However, the ability to generalize to unseen components remains unclear, as their data set was not split component-wise and their model was not disclosed to enable testing the model accordingly.

Lin et al. \cite{Lin2024} trained a directed message passing NN to synthetic vapor pressure data for more than 19,000 pure components, which were computed using the coefficients of the Wagner equation, the critical temperatures, and the critical pressures from the NIST-TRC data bank. The authors compared a direct prediction of vapor pressures with the prediction of parameters of embedded correlations for the vapor pressure curve. They report that, using a component-wise split, the Wagner equation yielded the best results on their test set, which is, however, to be expected when the Wagner equation is used for generating the data.

In this work, we propose \textbf{GRAPPA} - a \textbf{GRA}ph neural network for \textbf{P}redicting the \textbf{P}arameters of the \textbf{A}ntoine equation. In contrast to the two very recent works in which GNNs were used for vapor pressure prediction \cite{Santana2024, Lin2024}, and to which we compare our model, we use a graph attention network (GAT) \cite{Velickovic2018} for the message passing step, a pooling layer that can capture complex interactions through self-attention, and we train and evaluate exclusively on experimental data of 24,753 pure components. We hybridize our model by incorporating the Antoine equation in our prediction head, facilitating the straightforward implementation of GRAPPA into established process simulators, as it directly yields the Antoine parameters.

The paper is organized as follows: In the section "Data", the database, all data pre-processing steps, the data splitting, and the creation of the initial molecular graphs from SMILES are explained. The section "Methods" contains all model architecture and training information. In the "Results and Discussion" section, the model predictions are comprehensively evaluated and compared to literature methods. The links to the source code, the trained model, and our interactive website can be found in the section "Model Availability". Finally, in the "Conclusion", we summarize our findings and provide an outlook for future work.

\section{Data}
\subsection{Data Base and Curation}
\begin{figure}
    \begin{subfigure}{0.49\textwidth}
        \centering
        \includegraphics[width=\linewidth]{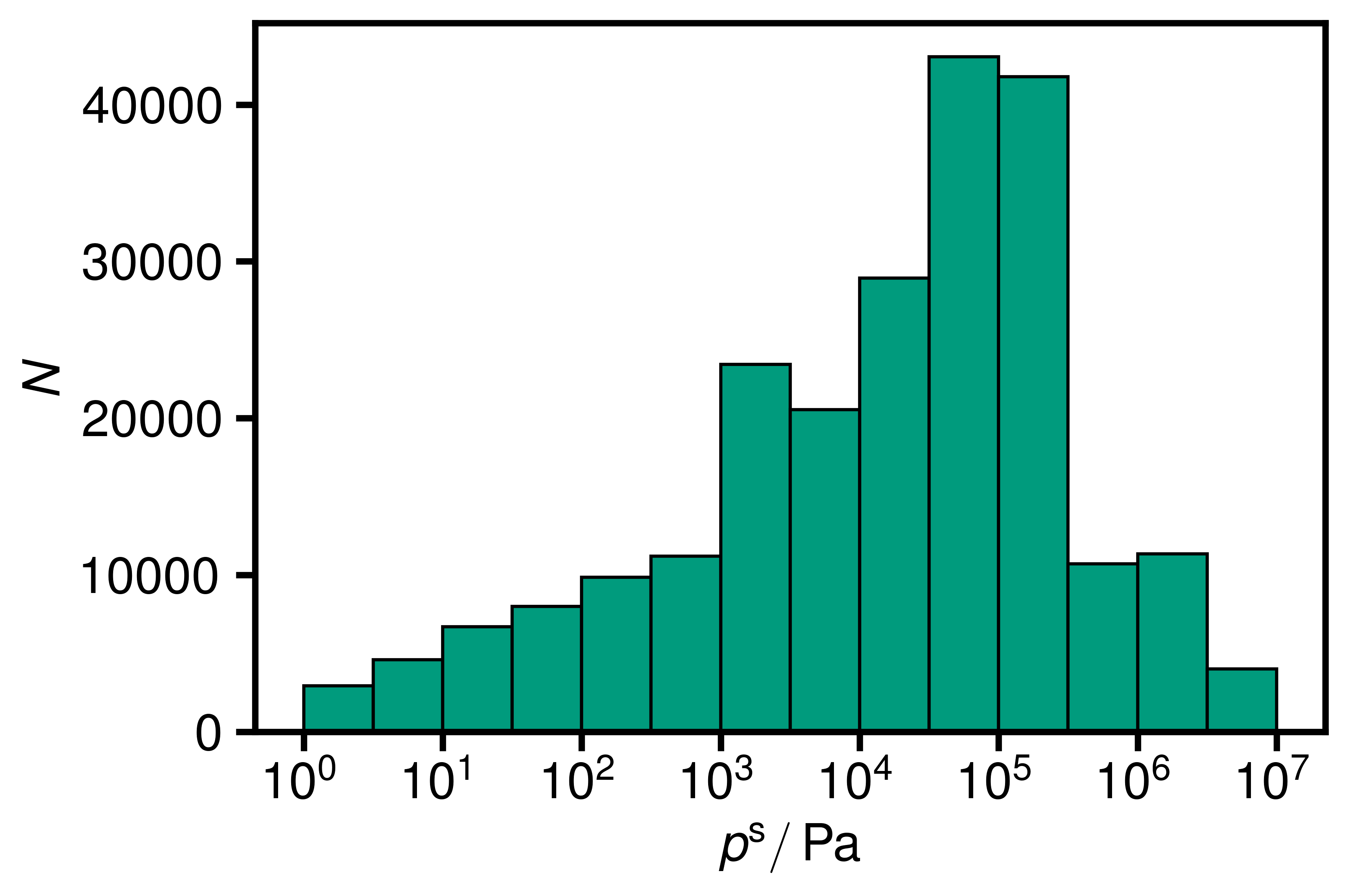}
    \end{subfigure}
    \hfill
    \begin{subfigure}{0.49\textwidth}
        \centering
        \includegraphics[width=\linewidth]{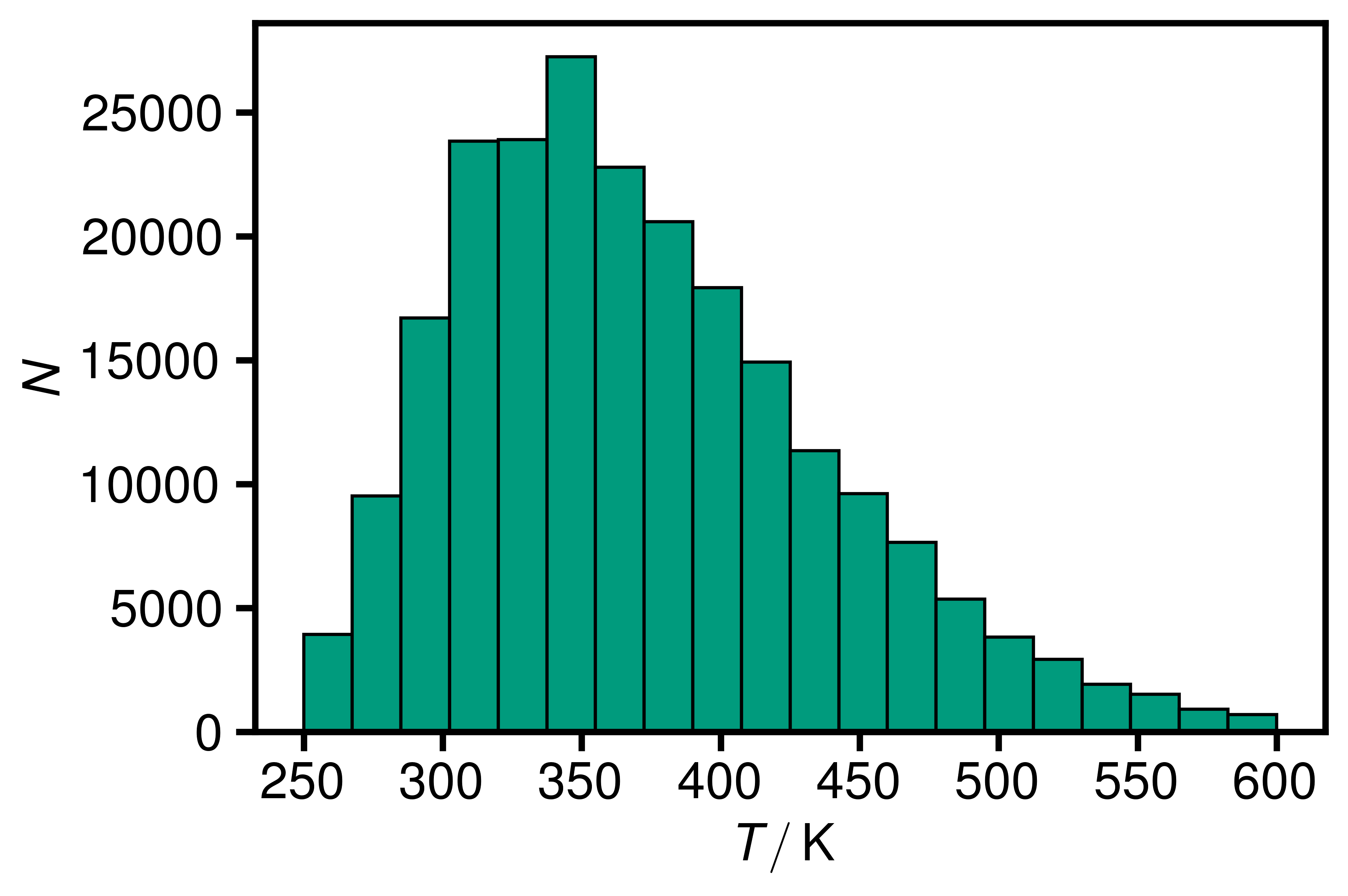}
    \end{subfigure}
    \caption{\label{fig:histograms_dataset}
    Histograms visualizing the distribution of the data points in the final data set after pre-processing. Left: Number of data points over the pressure. Right: Number of data points over the temperature.}
\end{figure}

All vapor pressure \ps data for pure components were retrieved from the 2024 version of the Dortmund Data Bank (DDB). \cite{DDB2024} We used only those components for which a canonical SMILES string could be generated from the DDB mol-files (describing the molecular structure) with the \textit{rdkit} \cite{rdkit} package. Each data point consists of a SMILES representing the molecule, the temperature, and the corresponding experimental value for \ps. For the training process of the GNN, sufficient training data for each node type (i.e., the atom the node represents) is required. Because these data are not available for components that contain less common atoms (such as Ge, Fe, ...), we have a priori restricted the chemical space to define a clear applicability range of the model. The same holds for the temperature and pressure ranges. Thus, in the pre-processing, we first only extracted data points that fulfilled the following criteria:
\begin{itemize}
    \item The component contains at least one carbon atom.
    \item The component only consists of the following atoms: C, N, O, Cl, S, F, Br, I, P.
    \item No atom in the component has formal charge or unpaired electrons.
    \item The temperature lies between 250 K and 600 K.
    \item The pressure lies between 1 and $10^{7}$ Pa.
\end{itemize}
Data points marked as "poor quality" in the DDB were also excluded. Furthermore, to enable the model to differentiate between stereoisomers, only those data points were kept for which the isomerism was correctly represented in the SMILES. In the next step, components with conflicting data points (e.g., two contradicting vapor pressure curves from different sources) were manually removed. To additionally eliminate single outliers, the Antoine equation (cf.~Eq.~(\ref{eq:antoine_equation})) was fitted (using robust least squares) individually for all components with at least five data points, and those data points with a relative deviation of \ps larger than $50\,\%$ from the Antoine fit were removed. For roughly a third of the components, there was only one available data point, and five or more data points were available for only about $17\,\%$ of the components. Consequently, removing faulty values or outliers was infeasible for a large share of the components. However, because it is crucial to cover the chemical space as well as possible, we decided to keep these components in the data set. After all pre-processing steps, the final data set contained 227,062 data points for 24,753 components. Fig.~\ref{fig:histograms_dataset} shows the distribution of the final data set over pressure and temperature in two histograms.

\subsection{Data Split}
Our final data set was split component-wise: $80\,\%$ of the components were used as the training set, $10\,\%$ as the validation set, and the remaining $10\,\%$ as the test set. This split was done randomly, except for data for molecules with less than five carbon atoms, which were all put in the training set. The motivation for this procedure is that the extrapolation from large to small molecules is difficult for GNNs due to their message passing steps (cf. "Model Architecture"). However, while smaller molecules are more difficult to model, they are usually well-measured, so prediction methods for their properties are less necessary than ones for more complex molecules, for which we test the model developed in this work. All data in the training set were used to fit the parameters of GRAPPA, and the validation set was used to optimize the hyperparameters and select the best model architecture. The test set was only used for the conclusive evaluation of GRAPPA. The final model disclosed with this work was trained to all available data.

\subsection{Generation of the Initial Molecular Graphs}
The SMILES of each molecule was converted to an initial molecular graph using \textit{rdkit} \cite{rdkit}. In the molecular graph $\mathcal{G}$, the atoms (except for hydrogen, which is treated implicitly, cf. below) are represented as a set of nodes $\mathcal{V}$, and the bonds are represented as a set of bonds $\mathcal{E}$. All nodes and edges are represented by an individual feature vector encoding information about the corresponding atom or bond. The vector containing the features of node $i$ is denoted as $\bm{x}_i$, and the features of the edge between nodes $i$ and $j$ are summarized in the vector $\bm{e}_{i,j}$. The node features in the initial molecular graph are the atom type, the number of bonds, the number of bonded hydrogen atoms, the hybridization, whether the atom is aromatic, and whether the atom is part of a ring. The edge features contain the bond type and whether the bond is conjugated, part of a ring, or part of a stereoisomer. All features with more than two classes were one-hot encoded, resulting in 24 node features and nine edge features. We have adapted some parts of the code of Sanchez-Medina et al.\cite{SanchezMedina2022} to implement the molecular graph generation.

\section{Methods}
\subsection{Model Architecture}
\begin{figure}
    \centering
    \includegraphics[width=\linewidth]{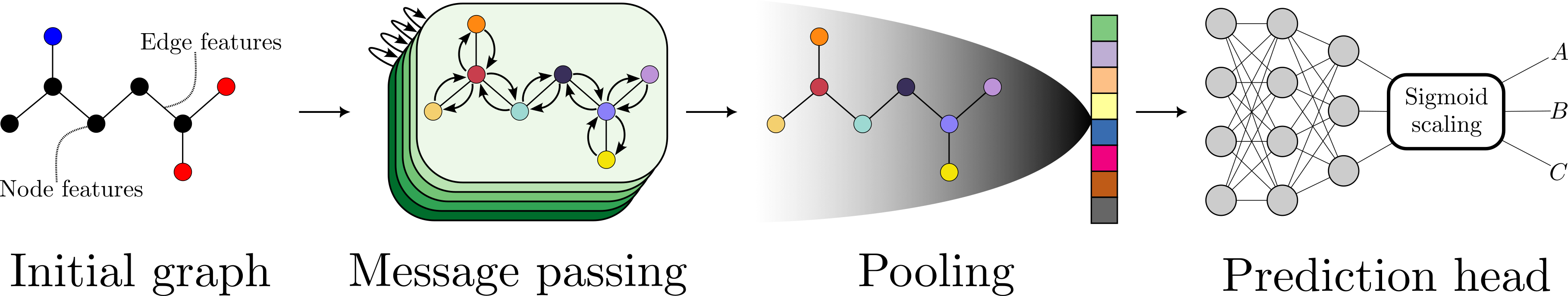}
    \caption{Schematic overview of the architecture of GRAPPA, with the initial molecular graph as the input and the three Antoine parameters ($A$,$B$,$C$) as the output. Details on the individual steps are given in the text.}
    \label{fig:model_overview}
\end{figure}

The proposed model consists of three parts: First, a message passing between the nodes is applied to the initial molecular graph to enrich the graph embedding with information about the vicinities of the nodes. Second, the pooling layer derives a numerical embedding of fixed length from the learned graph embedding. Finally, this numerical embedding serves as input for a FFNN that predicts component-specific vapor pressure curves in the form of Antoine parameters (subsequently called prediction head). The three parts are described in more detail in the following. A schematic overview of the model architecture is given in Fig.~\ref{fig:model_overview}.

\textbf{Message passing.} During the message passing, the graph embedding is updated, and information is exchanged between the nodes. From a chemical perspective, this enriches the node embeddings with information about their local environment and improves their expressivity. GRAPPA uses a graph attention network (GAT)\cite{Velickovic2018} for message passing, more specifically, the revised implementation \textit{GATv2Conv} \cite{Brody2022} in \textit{pytorch-geometric}. \cite{pytorch_geometric} The starting point of the message passing is the initial molecular graph to which multiple message passing layers are subsequently applied. In one message passing layer, the embeddings of all $N$ nodes in the graph are updated as follows:
\begin{equation}
\label{eq:gat_message_passing}
    \bm{x}_i^{'} = \sum_{j \in \mathcal{N}(i) \cup i} \alpha_{i,j}\bm{\Theta}_\mathcal{V}\bm{x}_j \quad \mathrm{for} \quad i=1,...,N
\end{equation}
Here, $\bm{x}_i^{'}$ is the updated embedding of node $i$, calculated based on the embeddings of node $i$ and all neighboring nodes $\mathcal{N}(i)$ from the previous layer, a trainable weight matrix $\bm{\Theta}_\mathcal{V}$, and the attention coefficient $\alpha_{i,j}$. In the first layer, the input embedding dimension (the length of the vector $\bm{x}_i$) is defined by the number of node features of the initial graph. For the output embedding of the first layer and all embeddings in the subsequent layers, we chose the embedding dimension $d=32$. The advantage of \textit{GATv2Conv} over other message passing layers is that the attention coefficient $\alpha_{i,j}$ is learnable and weighs the incoming information from the neighboring nodes differently. Thus, the local substructures that have a high impact on \ps can be identified by inspecting the values of $\alpha_{i,j}$ after model training; an example is given in section "Examples for Predicted Vapor Pressure Curves and Evaluation of Attention Scores". The attention coefficient for two neighboring nodes $i$ and $j$ is computed with
\begin{equation}
\label{eq:gat_attention_coefficient}
    \alpha_{i,j} = 
    \frac{ \exp{ \left( \bm{a}^\top \mathrm{LeakyReLU} \left( \bm{\Theta}_\mathcal{V}\bm{x}_i +  \bm{\Theta}_\mathcal{V}\bm{x}_j + \bm{\Theta}_\mathcal{E}\bm{e}_{i,j}\right) \right)}}
    {\sum_{k \in \mathcal{N}(i) \cup i} \exp{ \left( \bm{a}^\top \mathrm{LeakyReLU} \left( \bm{\Theta}_\mathcal{V}\bm{x}_i +  \bm{\Theta}_\mathcal{V}\bm{x}_j + \bm{\Theta}_\mathcal{E}\bm{e}_{i,k}\right) \right)}}
\end{equation}
Here, $\bm{a}$ and $\bm{\Theta}_\mathcal{E}$ are trainable weight matrices, and \textit{LeakyReLU} is the leaky rectified linear unit activation function. The embedding vector of the edge between nodes $i$ and $j$ is denoted as $\bm{e}_{i,j}$. We have used the multi-head attention option of the \textit{GATv2Conv} module. The idea behind multi-head attention is to employ Eqs.~(\ref{eq:gat_message_passing}) and~(\ref{eq:gat_attention_coefficient}) multiple times with different, independent parametrizations. Each of these attention heads thereby learns to focus on other substructures. In a multi-head attention with $N_{\mathrm{MH}}$ attention heads, the message passing layer thus produces $N_{\mathrm{MH}}$ distinct updated node embeddings $\bm{x}_i^{'}$ for each node. The mean of the $N_{\mathrm{MH}}$ node embeddings is then calculated to retain the original embedding size.

\textbf{Pooling function.} The pooling function takes the final graph embedding $\bm{X} \in \mathbb{R}^{N \times d}$ and produces a fixed-size numerical embedding $\bm{h} \in \mathbb{R}^{d}$ for each molecule. This step is necessary as the molecules (and therefore the graphs) differ in size, but the prediction head requires an embedding of a pre-defined size as input. In our work, we tested two pooling variants. The first one is standard \textit{sum} pooling where the fixed-size embedding $\bm{h}$ is calculated as sum of all individual node embeddings $\bm{x}_i$ in the graph:
\begin{equation}
            \bm{h} = \sum_{i=1}^{N} \bm{x}_i
\label{eq:sum_pooling}
\end{equation}
Additionally, we have tested an attention-based pooling method, similar to Refs.~\cite{Buterez2023, Baek2021}, which is built upon the self-attention algorithm \cite{Vaswani2017}. In the first step, using the trainable weight matrices $\bm{W}_\mathrm{q} \in \mathbb{R}^{d \times k}, \bm{W}_\mathrm{k}  \in \mathbb{R}^{d \times k}$ and $\bm{W}_\mathrm{v} \in \mathbb{R}^{d \times d}$, the query matrix $\bm{Q}$, the key matrix $\bm{K}$, and the value matrix $\bm{V}$ are built from the node embedding matrix $\bm{X}$:
\begin{equation}
        \bm{Q} = \bm{X} \cdot \bm{W}_\mathrm{q}, \quad \bm{K} = \bm{X} \cdot \bm{W}_\mathrm{k}, \quad \bm{V} = \bm{X} \cdot \bm{W}_\mathrm{v} \quad \mathrm{with} \quad \bm{Q}, \bm{K} \in \mathbb{R}^{N \times k};\quad \bm{V} \in \mathbb{R}^{N \times d}
        \label{eq:interaction_pooling_QKV_matrices}
\end{equation}
For simplicity, we have chosen the key dimension $k$ equal to the embedding dimension ($k=d=32$). The three matrices are used to calculate the context matrix
\begin{equation}
    \bm{Z} = \mathrm{softmax}\left(\frac{\bm{Q}\bm{K}^\top}{\sqrt{d}}\right) \cdot \bm{V} \quad \mathrm{with} \quad \bm{Z} \in \mathbb{R}^{N \times d}.
    \label{eq:interaction_pooling_context_matrix}
\end{equation}
The first term $\mathrm{softmax}(...)$ is a matrix of dimension $\mathbb{R}^{N \times N}$ and represents the attention weights, i.e. values for the interaction of all binary node pairs $i$ and $j$ in the graph. The multiplication with the value matrix $\bm{V}$ results in the context matrix $\bm{Z}$, which has the same dimensions as the graph embedding matrix $\bm{X}$ after the message passing, but contains information on the interaction between all nodes in the graph. Finally, the molecule's fixed-size embedding $\bm{h}$ is obtained by summing up the context vectors $\bm{z}_i$ of all nodes in the graph, analogous to Eq.~(\ref{eq:sum_pooling}). This pooling approach allows the model to capture interactions between all atoms in the molecules, which is especially relevant for large molecules with multiple functional groups farther apart than the message passing distance. It can also be interpreted as a final, global message passing step, followed by \textit{sum pooling}. We refer to this approach as \textit{interaction pooling} in the following.  

\textbf{Prediction head.} The prediction head predicts the Antoine parameters using the numerical embedding $\bm{h}$ from the pooling function and the number of hydrogen donors and acceptors in the molecule, which were obtained through \textit{rdkit} \cite{rdkit}. These are concatenated to the embedding $\bm{h}$ because preliminary studies showed that this increases the prediction accuracy. The prediction head was realized by a feed-forward neural network (FFNN) using hidden layers with 16 neurons. We employed batch normalization and the \textit{ELU} activation function between all layers. The last layer of the FFNN yields three outputs for the three Antoine parameters. To obtain the final parameters $A, B$, and $C$ (cf.~Eq.~(\ref{eq:antoine_equation})), they are scaled using sigmoid functions. We found this approach to increase training stability and decrease the number of epochs to convergence. The parameter ranges for the scaling were defined as follows:  
\begin{equation}
\label{eq:antoine_param_ranges}
    A \in [5,20], \quad B \in [1500, 6000], \quad C \in [-300,0].
\end{equation}
These were determined by fitting the Antoine equation for all components with at least ten data points in the training set. The parameter ranges in Eq.~(\ref{eq:antoine_param_ranges}) were chosen slightly larger than those obtained by the fit to allow for extrapolation to unseen components with different behavior than those in the training set. By limiting the range of the \textit{B} parameter to positive values, we have enforced the correct slope of the vapor pressure curve. The final prediction for the vapor pressure is then calculated using Eq.~(\ref{eq:antoine_equation}) and the temperature in Kelvin.

\subsection{Model Training}
The model is implemented using \textit{pytorch} \cite{pytorch} and \textit{pytorch-geometric} \cite{pytorch_geometric}, and the message passing layer, the pooling function, and the prediction head (including the Antoine equation) are connected in the computational graph and can therefore be trained in an end-to-end manner. The model training was divided into two parts: First, we performed a pre-training with the mean squared error (MSE, cf.~Eq.(\ref{eq:mae_mse})) as the loss function for 100 epochs. Then, we trained for 200 epochs using the Huber loss with a threshold of 0.5. The Huber loss transitions from the MSE loss to the mean absolute error (MAE, cf.~Eq.(\ref{eq:mae_mse}) loss at the threshold, which prevents outliers from having a too large impact on the loss and, therefore, on the adjustment of the parameters. In both trainings, we used the \textit{AdamW} optimizer \cite{Loshchilov2017}. We used the \textit{OneCycleLR} \cite{Smith2017} learning rate scheduler for the pre-training and the \textit{ReduceLROnPlateau} scheduler for the main training. To prevent overfitting, an early stopping strategy was applied, whereby the best model on the validation set based on the median absolute percentage deviation ($\mathrm{MAPE}_i$, cf. "Training and Evaluation Metrics") was chosen. 

We performed a grid search over the number of GAT layers, the number of GAT attention heads, and the number of hidden layers in the prediction head. Furthermore, we tested standard \textit{sum} pooling and the \textit{interaction pooling} approach presented above. The best model in this grid search had four GAT layers with two GAT attention heads each, three hidden layers in the prediction head, and used \textit{interaction pooling}. All hyperparameters and the ranges covered in the grid search are given in Tab.~S1 of the Supporting Information. The final model contains 15,319 trainable parameters.

\subsection{Training and Evaluation Metrics}
Different error metrics were used for the training and evaluation of GRAPPA. The mean absolute error (MAE) and mean squared error (MSE) 
\begin{equation}
\begin{aligned}
\label{eq:mae_mse}
\mathrm{MAE} &= \frac{1}{M}\sum_{i=1}^{M}\vert \ln\left(p^\mathrm{s}_{\mathrm{pred}, i} / \mathrm{kPa}\right) - \ln\left(p^\mathrm{s}_{\mathrm{exp}, i} / \mathrm{kPa}\right) \vert \\
\mathrm{MSE} &= \frac{1}{M}\sum_{i=1}^{M}\left( \ln\left(p^\mathrm{s}_{\mathrm{pred}, i} / \mathrm{kPa}\right) - \ln\left(p^\mathrm{s}_{\mathrm{exp}, i} / \mathrm{kPa}\right) \right)^2
\end{aligned}
\end{equation}
were calculated based on the natural logarithm of the vapor pressures in kPa. Here, $M$ denotes the number of data points in the considered data set. Because it is a more practical error metric, we also use absolute percentage error (APE) of each data point \textit{i} defined as
\begin{equation}
    \mathrm{APE}_i = \left| \frac{p^\mathrm{s}_{\mathrm{pred}, i}  - p^\mathrm{s}_{\mathrm{exp}, i}}{p^\mathrm{s}_{\mathrm{exp},i}}\right|  \cdot 100\,\%.
    \label{eq:ape_i}
\end{equation}
Additionally, we calculated the component-wise absolute percentage error $\mathrm{APE}_C$ by averaging the $\mathrm{APE}_i$ over the $K$ data points of the considered component $C$:
\begin{equation}
        \mathrm{APE}_C = \frac{1}{K}\sum_{i=1}^{K} \mathrm{APE}_i
        \label{eq:ape_C}
\end{equation}
To measure the prediction accuracy over a considered data set, besides MAE and MSE, the medians of the $\mathrm{APE}_i$ and $\mathrm{APE}_C$ were used, as the median is more robust towards outliers in the data than the mean; they are labeled $\mathrm{MAPE}_i$ and $\mathrm{MAPE}_C$, respectively, in the following. Because the $\mathrm{MAPE}_C$ is less biased towards well-studied components, it is used as the primary evaluation metric.

\section{Results and Discussion}
\subsection{Performance of GRAPPA for Vapor Pressure Prediction}
Tab.~\ref{tab:model_evaluation} summarizes all GRAPPA evaluation metrics on the train, validation, and test set. 
\begin{table}[]
    \centering
    \caption{Error scores of GRAPPA for modeling \ps from the training (train), validation (valid), and test data set. The $\mathrm{MAPE}_C$ is given for all components in the respective set, for components with at least two data points ($K\geq2$) in the set only, and for components with at least five data points ($K\geq5$) in the set only.}
    \begin{tabular}{lcccccc}
         \hline \hline
         Data set &  MAE & MSE & $\mathrm{MAPE}_i$ & $\mathrm{MAPE}_C$ & $\mathrm{MAPE}_C$ ($K\geq2$)  & $\mathrm{MAPE}_C$ ($K\geq5$)\\
         \hline
         Train & 0.127 & 0.148 & 4.39& 22.77\,\% & 13.65\,\% & 9.60\,\%   \\
         Valid & 0.233 & 0.480 & 8.53 & 26.90\,\% & 17.45\,\% & 12.35\,\%  \\
         Test  & 0.201 & 0.301 & 6.33 & 26.70\,\% & 16.42\,\% & 12.54\,\%  \\
         \hline \hline
    \end{tabular}
    \label{tab:model_evaluation}
\end{table}

The scores are of similar magnitude for all data sets, demonstrating that the model did not overfit the training data and can generalize well to unseen components. The table lists the $\mathrm{MAPE}_C$ for all components and, separately, only for those with at least two or at least five experimental data points. The $\mathrm{MAPE}_C$ for components with at least two data points is almost half that for all components for all data sets, suggesting that the model predictions for the single-measured components deviate stronger from the experimental values. As discussed above (cf. "Data Split"), no checks regarding faulty data or outliers could be carried out for components with less than five data points. Because these larger deviations for single-measured components appear not only in the test data set but also in the training and validation data set, we can assume that many of these single data points carry significant measurement errors and are poorly suited for model evaluation. Because these data potentially skew the $\mathrm{MAPE}_C$, further evaluation is carried out only on the components for which at least two experimental data points are available. More information on the influence of the number of experimental data points on the prediction accuracy in the test set is given in Fig.~S6 of the Supporting Information.

\begin{figure}
    \centering
    \includegraphics[width=0.8\linewidth]{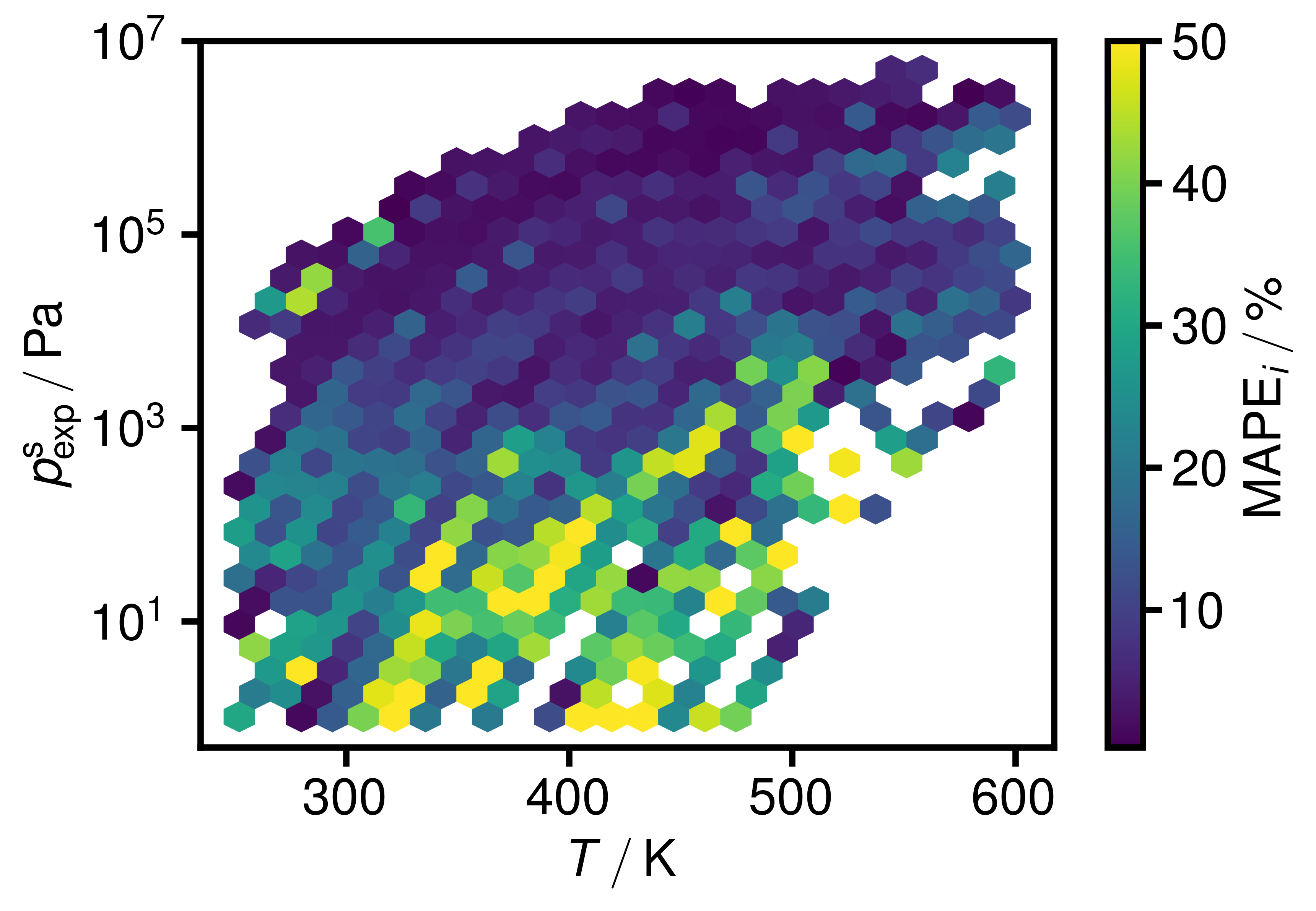}
    \caption{Hexbin plot to visualize the relation between the $\mathrm{MAPE}_i$, the pressure, and the temperature on the test data set. Only results for components with at least two experimental data points are shown. Each hexagon is colored according to the $\mathrm{MAPE}_i$ of all data points in the covered area. Values larger than $50\,\%$ are clipped for readability.}
    \label{fig:hexplot_pressure_temperature}
\end{figure}

In Fig.~\ref{fig:hexplot_pressure_temperature}, a hexbin plot in terms of the $\mathrm{MAPE}_i$ on the test data is shown to visualize the dependence of the prediction accuracy on pressure and temperature. 
A significant decline in prediction accuracy can be observed for very small pressures, approximately below 1~kPa. There are three reasons for this: First, and most importantly, the measurement uncertainties in the low-pressure regime are much larger than at higher pressures. Second, using a single set of Antoine parameters to describe the entire vapor pressure curve over a wide pressure range is known to cause issues in either the low- or the high-pressure regime. Third, the number of training data points in the low-pressure regime is smaller than in the medium- to the higher-pressure regime (cf.~Fig.~\ref{fig:histograms_dataset}), leading to a generally lower prediction accuracy for low pressures. Given these results, the prediction of GRAPPA for pressures below 1~kPa should be treated carefully. In contrast to pressure, temperature does not significantly impact prediction accuracy. The reader is referred to the Supporting Information for a deeper analysis of the pressure and temperature dependence of the prediction accuracy in Figs.~S3 and S4.

\subsection{Comparison to Literature Methods}
\begin{figure}[!h]
    \centering
    \includegraphics[width=0.8\linewidth]{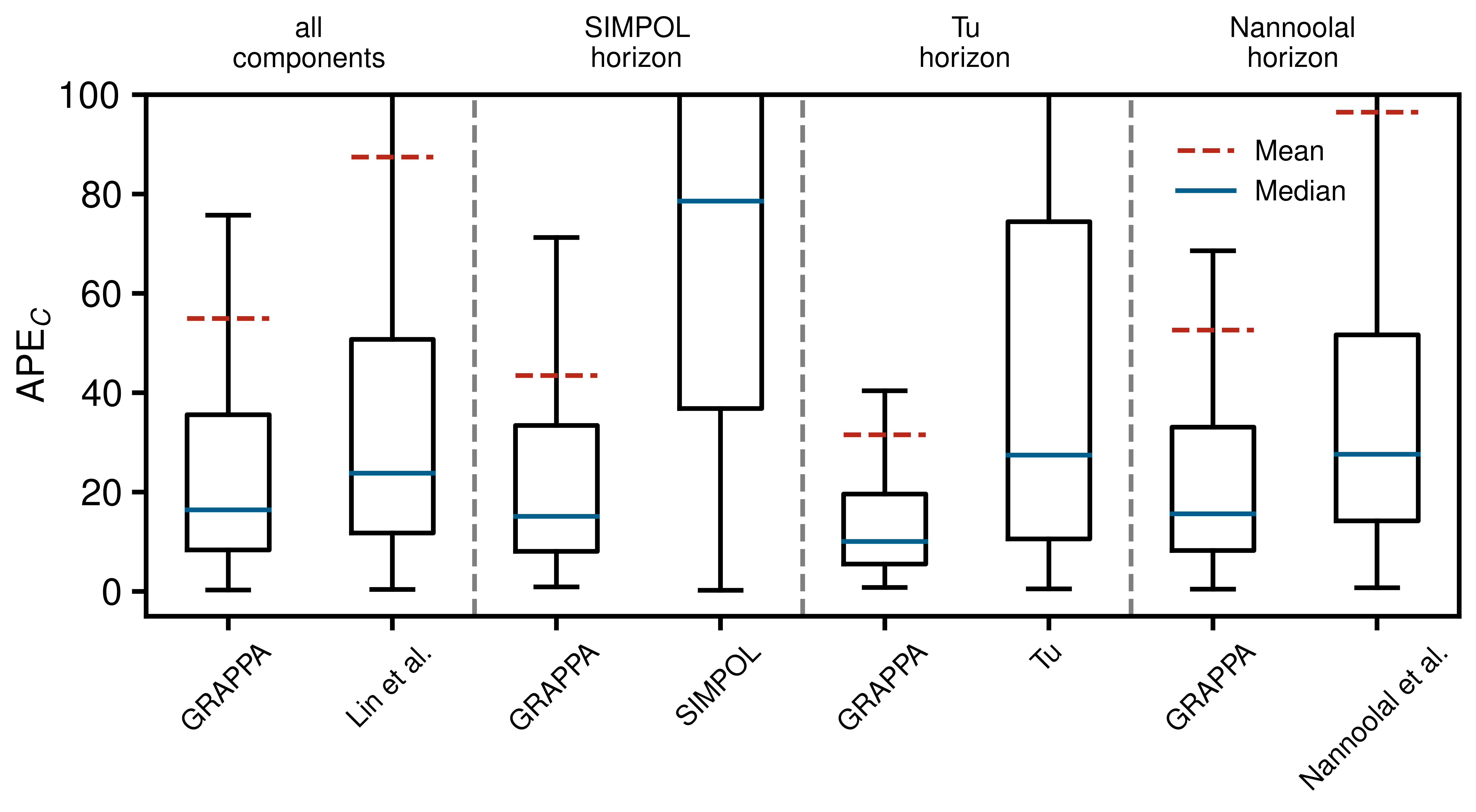}
    \caption{Boxplot comparing the prediction accuracy of GRAPPA with four literature methods on the test set of GRAPPA for components with at least two experimental data points. The boxes represent the interquartile range, and the whiskers are 1.5 times the interquartile range. Because some literature methods are limited in their applicability, GRAPPA was also evaluated only on components in their scope: the SIMPOL \cite{Pankow2008} horizon contains $53.8\,\%$ of our test components, that of the method by Tu \cite{Tu1994} $36.1\,\%$, and that of the method by Nannoolal et al. \cite{Nannoolal2008} $94.9\%$. The method by Lin et al. \cite{Lin2024} is applicable to our entire test set.}
    \label{fig:boxplot}
\end{figure}

We directly benchmarked GRAPPA against three group contribution methods from the literature for which the parametrization is available, namely the methods by Tu \cite{Tu1994}, the SIMPOL method \cite{Pankow2008}, and the method by Nannoolal et al. \cite{Nannoolal2008} (incorporating the corresponding normal boiling point prediction method \cite{Nannoolal2004} from the same authors). These methods were chosen for comparison because they require only the molecular structure as input and are, therefore, comparable to GRAPPA in terms of general applicability. However, due to missing parameters for certain structural groups, the three group contribution methods are only applicable to a subset of the test data; more specifically, SIMPOL is only applicable to $53.8\,\%$ of the components from the test set (SIMPOL horizon), the method by Tu to $36.1\,\%$ (Tu horizon), and the method by Nannolal et al.~to $94.9\%$ (Nannolal horizon). GRAPPA was also evaluated separately on these subsets of our test set to enable a fair comparison. The results of this comparison in terms of $\mathrm{APE}_C$ on the test set from this work are shown in a boxplot in Fig.~\ref{fig:boxplot}. It should be noted that this comparison is likely biased in favor of the literature methods, as it can be assumed that at least some of the data in our test set were used for training these methods, while this was not the case for GRAPPA. This is especially relevant for the method by Nannoolal et al., which was fitted to large amounts of data from the DDB, almost certainly including a large share of the components from our test set. Despite this bias in favor of the literature methods, GRAPPA significantly outperforms all three group contribution methods in both applicability and accuracy. 

Of the three EoS-based prediction methods for vapor pressures discussed above, only Felton et al. \cite{Felton2024} disclosed their model. However, because they only included molecules with less than 15 atoms in their training data and our test data set contains mainly larger molecules, we refrained from a quantitative comparison on our test data set. On their test data set, they state an average error of $40\,\%$ for the prediction of vapor pressures, which is much higher than the error scores of GRAPPA. Habicht et al. \cite{Habicht2023} do not give an explicit number for their prediction accuracy for the vapor pressure but state that in most cases, the average absolute relative deviation (AARD, analogous to our $\mathrm{APE}_C$) is below $25\,\%$. The best scores of the EoS-based models are reported by the authors of SPT-PC-SAFT \cite{Winter2023} with an $\mathrm{MAPE}_C$ of $8.7\,\%$. However, their data set has been extensively cleaned and contains only components with at least five data points for the vapor pressure and at least three data points for the density, which were also used for training their model.

Finally, we compare the results of GRAPPA to the two recent works that also use GNNs for the prediction of pure component vapor pressures. The trained PUFFIN \cite{Santana2024} model is not disclosed, preventing a comparison on the same test data. The authors report an MSE of 0.1609 on the base-10 logarithmic vapor pressures on their test set. Converted to the natural logarithmic vapor pressures, this corresponds to an MSE of 0.853, which is significantly higher than the MSE of GRAPPA on our test set, which is 0.301.\footnote[3]{The MAE and MSE of GRAPPA reported here were calculated on the entire test data set, including the components with only a single experimental data point, of which we assume many to be erroneous.} Lin et al. \cite{Lin2024} disclosed their trained model, which we evaluated on our test set, thereby ignoring that some of our test data might have been part of their training set (cf. discussion on the favorable bias above). We used their best model, more precisely, the published ensemble of ten models with the embedded Wagner equation, and provided the canonical SMILES and the temperature as input. The results are also included in the boxplot in Fig.~\ref{fig:boxplot} (left). Despite comparing a single GRAPPA model with an ensemble of their models (of which each has 20 times more parameters than GRAPPA), GRAPPA significantly outperforms the model of Lin et al. Besides demonstrating the high prediction accuracy of GRAPPA, these results also indicate that the Wagner equation, as used in the model of Lin et al., while being indisputably the more accurate correlation compared to the Antoine equation when fitted, simply because of the increased flexibility due to the higher number of parameters, is not necessarily the better choice in a predictive scenario.

\subsection{Boiling Point Prediction}
The Antoine equation can easily be rearranged to allow predictions for the boiling temperature at a given pressure. Hence, the Antoine parameters predicted with GRAPPA can directly be used for predicting boiling temperatures for any component whose molecular structure is known and that fulfills the criteria defined for the pre-processing (cf.~section "Data Preparation"). Because the normal boiling point $T_\mathrm{b}$ is of great technical relevance, we evaluate the prediction accuracy of GRAPPA for $T_\mathrm{b}$ in the following. 

For this purpose, we have first collected the experimental data points from our test set with a pressure between 99 kPa and 102 kPa (we, again, restrict the study to components for which at least two data points, irrespective of the pressure, were in our test set). In cases with more than one data point in that range, we calculated the mean pressure and temperature. Using the Antoine parameters obtained with GRAPPA and the (mean) pressures, we predicted $T_\mathrm{b}$ for each component. 

\begin{figure}
    \centering
    \begin{subfigure}{0.49\textwidth}
        \centering
         \includegraphics[width=\linewidth]{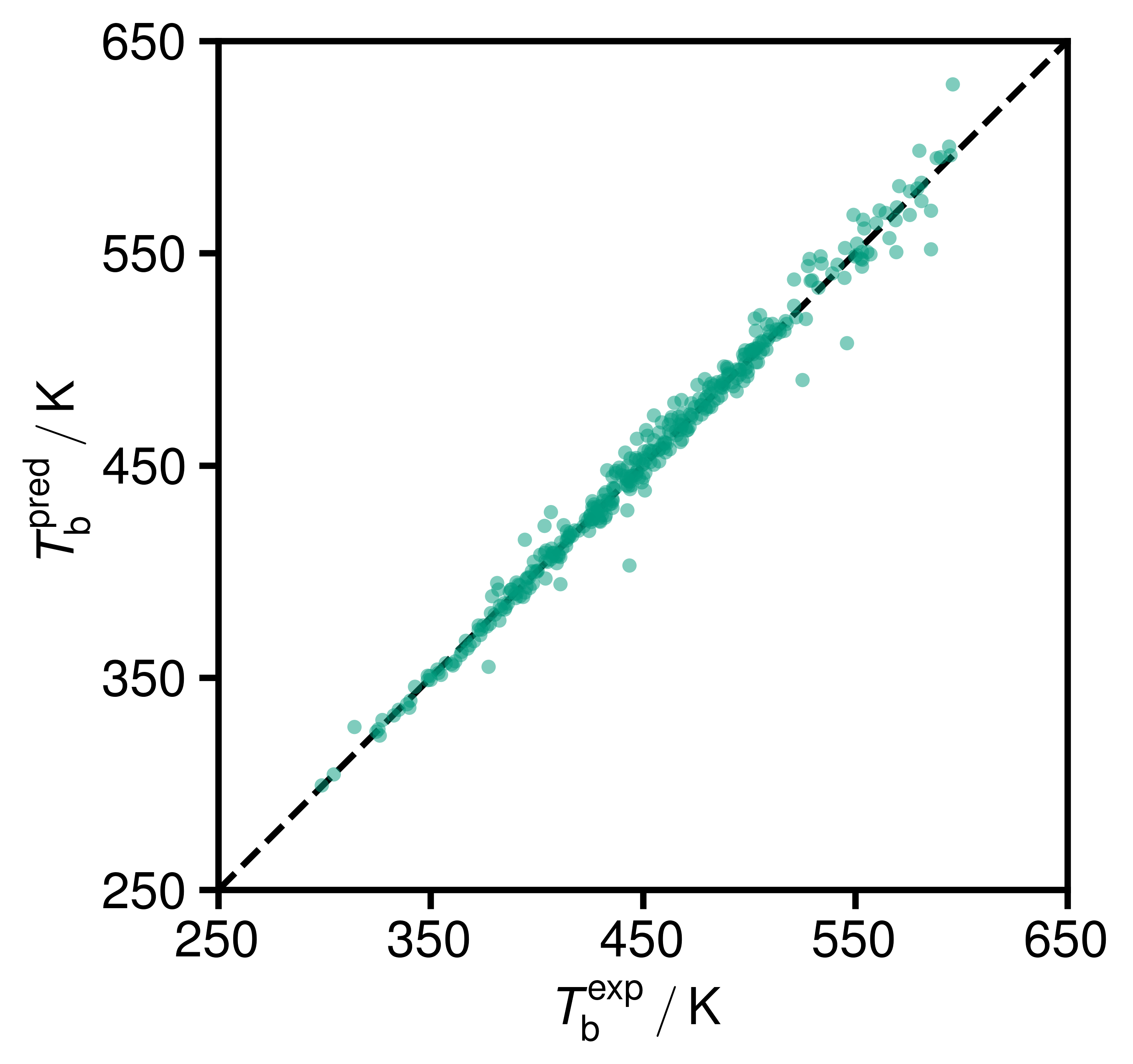}
    \end{subfigure}
    \hfill
    \begin{subfigure}{0.49\textwidth}
        \centering
         \includegraphics[width=\linewidth]{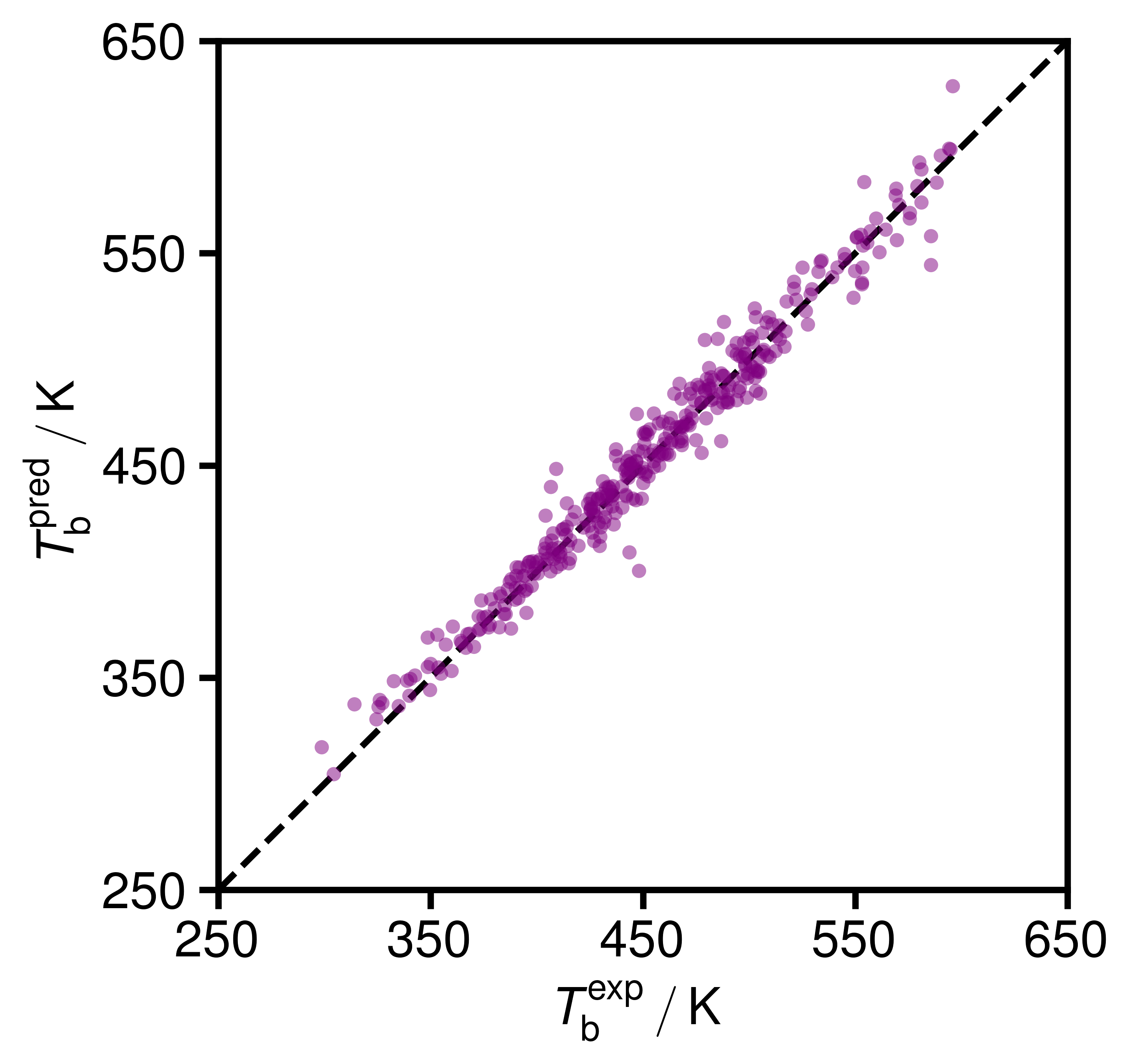}
    \end{subfigure}
    \caption{Parity plots showing predicted normal boiling points with GRAPPA~(left) and the group contribution method by Nannoolal et al. \cite{Nannoolal2004}~(right) over our experimental test data. Only data for components with at least two experimental data points in our test set are shown. For twelve of the components, a prediction using the method by Nannoolal et al. was infeasible due to problems with group fragmentation. The dashed line marks perfect predictions. }    \label{fig:parity_plot_nbp}
\end{figure}

A comparison of the GRAPPA predictions and the experimental data for the normal boiling point is shown in a parity plot in the left panel of Fig.~\ref{fig:parity_plot_nbp}. The mean absolute error of $T_\mathrm{b}$ is $4.76\,\mathrm{K}$ and the mean relative error is $1.05\,\%$. For comparison, we also calculated the normal boiling points of the same components with the GCM of Nannoolal et al. \cite{Nannoolal2004} and show the corresponding parity plots in the right panel of Fig.~\ref{fig:parity_plot_nbp}. Only for twelve components, the method by Nannoolal et al. could not be applied. The mean absolute error is $7.88\,\mathrm{K}$ and the mean relative error is $1.73\,\%$ and hence, clearly larger than for GRAPPA, despite the bias in favor of the method of Nannooal, which has presumably been trained on most of these data points, cf. discussion above. 

A recent publication \cite{Qu2022} of a GNN that was explicitly trained for predicting normal boiling points achieves a similar performance as GRAPPA with a mean absolute error of $5.78\,\mathrm{K}$ and a mean relative error of $1.32\,\%$ on their test set, which is comparable in its size (385 components (theirs) vs. 367 components (ours)). These results prove that GRAPPA is also a powerful tool for predicting (normal) boiling temperatures, even though it has not been explicitly trained for this task.

\subsection{Examples for Predicted Vapor Pressure Curves and Evaluation of Attention Scores }
\begin{figure}
    \centering
    \includegraphics[width=\linewidth]{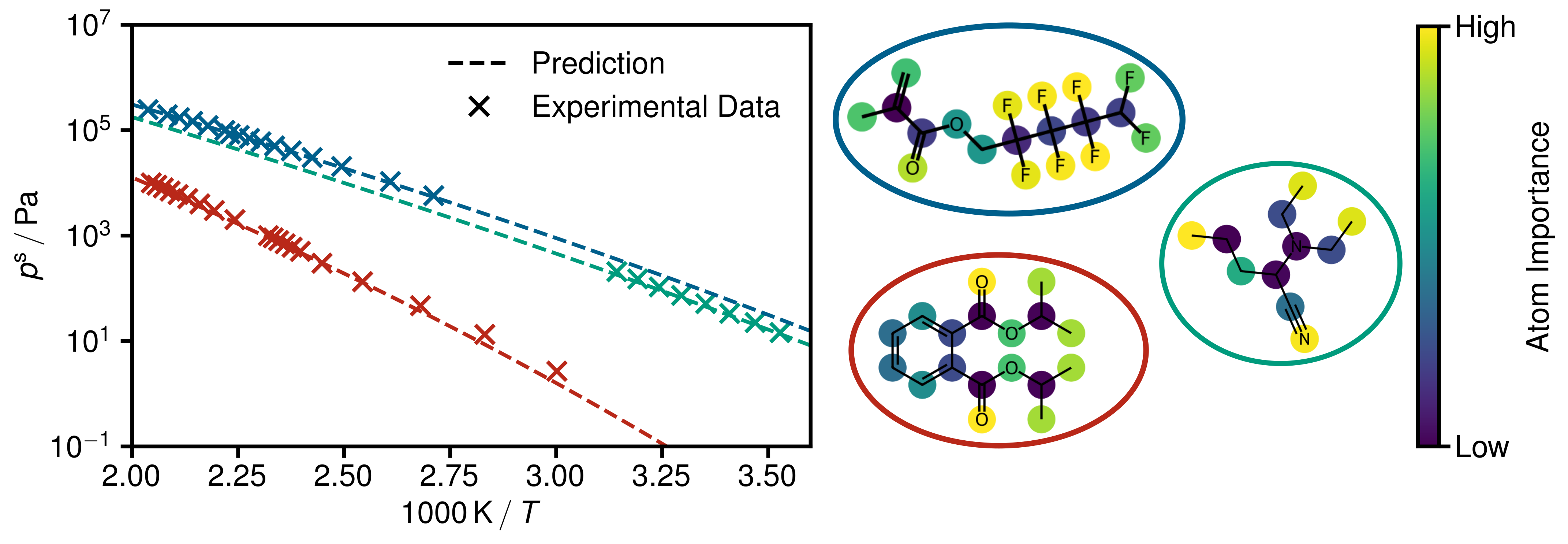}
    \caption{GRAPPA predictions and experimental data for the vapor pressure of three multi-functional components from our test set. The corresponding molecules are depicted on the right-hand side (CAS-RN from top to bottom: 355-93-1, 19340-91-1, 605-45-8). Each atom is colored based on its importance for the GRAPPA predictions, calculated by the mean of its outgoing attention coefficients in the last GAT layer. The scores are normalized for every molecule.}
    \label{fig:complex_molecules_curves}
\end{figure}

GRAPPA profits from the combination of its GNN architecture with the attention-based pooling function, which allows it to capture the effects of complex interactions in multi-functional components. In Fig.~\ref{fig:complex_molecules_curves}, we show vapor pressure curves for three such components as predicted with GRAPPA compared to experimental test data. 
The results show GRAPPA's excellent predictive accuracy, even in challenging cases. Next to the plot, the molecules are visualized, and their atoms are colored based on the mean of their outgoing attention coefficients in the last GAT layer, cf. Eq.~(\ref{eq:gat_attention_coefficient}). The color reflects each atom's importance in vapor pressure prediction. Notably, fluorine atoms, carbonyl groups, and open ends of alkane chains seem to play a prominent role in the prediction, which is in line with chemical intuition. The results show that the predictions of GRAPPA are not only accurate but also interpretable.

\section{Conclusions}
This work introduces GRAPPA - a broadly applicable and fully disclosed machine learning model for the prediction of pure component vapor pressures. GRAPPA is based on a graph neural network architecture and requires only the molecular structure as input. We have trained GRAPPA on more than 200,000 experimental data points of more than 20,000 pure components and evaluated it on a test set of unseen components, showing high prediction accuracy. GRAPPA predicts component-specific parameters of the Antoine equation, making the implementation in industrial applications, e.g., in process simulators, straightforward. Furthermore, GRAPPA can be used directly to predict boiling temperatures at a given pressure. 

Despite the broad applicability of GRAPPA, the following limitations must be considered. As shown in the results, the prediction accuracy declines for pressures below $1\,\mathrm{kPa}$, which may be due to the inability of the Antoine equation to describe the entire vapor pressure curve and the decline of measurement accuracy for low pressures. Therefore, GRAPPA should be used with care in such scenarios, e.g., for very large molecules. The chemical space in which GRAPPA can be used is limited to components that fulfill the criteria in our pre-processing. However, this covers most components of technical relevance in the chemical industry.

We have shown that GRAPPA performs significantly better than group contribution methods, which are the current state of the art for predicting vapor pressures or boiling temperatures in industry. Furthermore, a direct comparison was made between GRAPPA and another GNN based prediction method. The results demonstrate GRAPPA's higher prediction accuracy despite its reduced complexity, both regarding the underlying equation describing the vapor pressure (Antoine equation vs. Wagner equation) and the number of parameters. Future modifications of GRAPPA might use information on the critical point and related properties (e.g., the enthalpy of vaporization) in the training to exploit thermodynamic relations. Additionally, the uncertainties in both the experimental data and the predictions could be taken into account to enhance confidence in the model.

\section{Model Availability}
Our github repository \texttt{github.com/marco-hoffmann/GRAPPA} contains the final model trained on all data. The repository includes an example code that explains how to calculate Antoine parameters, vapor pressures, and (normal) boiling temperatures with GRAPPA.
Moreover, we have published an online prediction tool via our website \texttt{ml-prop.mv.rptu.de}, where the user can obtain predictions from GRAPPA without having to download or install the required code.

\begin{acknowledgement}
We gratefully acknowledge financial support by the Carl Zeiss Foundation in the frame of the project "Process Engineering 4.0" and by DFG in the frame of the Priority Program SPP2363 "Molecular Machine Learning" (grant number 497201843). Furthermore, FJ gratefully acknowledges financial support by DFG in the frame of the Emmy-Noether program (grant number 528649696). We would like to thank Jürgen Rarey for fruitful discussions regarding this work.

\end{acknowledgement}

\begin{suppinfo}
We provide a Supporting Information (SI), which provides details on hyperparameters of the model and the grid search. Additionally, the SI includes a more detailed analysis of the prediction accuracy of GRAPPA.
\end{suppinfo}

\providecommand{\latin}[1]{#1}
\makeatletter
\providecommand{\doi}
  {\begingroup\let\do\@makeother\dospecials
  \catcode`\{=1 \catcode`\}=2 \doi@aux}
\providecommand{\doi@aux}[1]{\endgroup\texttt{#1}}
\makeatother
\providecommand*\mcitethebibliography{\thebibliography}
\csname @ifundefined\endcsname{endmcitethebibliography}  {\let\endmcitethebibliography\endthebibliography}{}


\begin{mcitethebibliography}{64}
\providecommand*\natexlab[1]{#1}
\providecommand*\mciteSetBstSublistMode[1]{}
\providecommand*\mciteSetBstMaxWidthForm[2]{}
\providecommand*\mciteBstWouldAddEndPuncttrue
  {\def\EndOfBibitem{\unskip.}}
\providecommand*\mciteBstWouldAddEndPunctfalse
  {\let\EndOfBibitem\relax}
\providecommand*\mciteSetBstMidEndSepPunct[3]{}
\providecommand*\mciteSetBstSublistLabelBeginEnd[3]{}
\providecommand*\EndOfBibitem{}
\mciteSetBstSublistMode{f}
\mciteSetBstMaxWidthForm{subitem}{(\alph{mcitesubitemcount})}
\mciteSetBstSublistLabelBeginEnd
  {\mcitemaxwidthsubitemform\space}
  {\relax}
  {\relax}

\bibitem[DDB(2024)]{DDB2024}
Dortmund Data Bank. www.ddbst.com, 2024\relax
\mciteBstWouldAddEndPuncttrue
\mciteSetBstMidEndSepPunct{\mcitedefaultmidpunct}
{\mcitedefaultendpunct}{\mcitedefaultseppunct}\relax
\EndOfBibitem
\bibitem[Yaws(2015)]{Yaws2015}
Yaws,~C.~L. \emph{The Yaws handbook of vapor pressure}, second edition ed.; Gulf Professional Publishing is an imprint of Elsevier: Kidlington, Oxford, 2015\relax
\mciteBstWouldAddEndPuncttrue
\mciteSetBstMidEndSepPunct{\mcitedefaultmidpunct}
{\mcitedefaultendpunct}{\mcitedefaultseppunct}\relax
\EndOfBibitem
\bibitem[Lee and Kesler(1975)Lee, and Kesler]{Lee1975}
Lee,~B.~I.; Kesler,~M.~G. A generalized thermodynamic correlation based on three‐parameter corresponding states. \emph{AIChE Journal} \textbf{1975}, \emph{21}, 510--527\relax
\mciteBstWouldAddEndPuncttrue
\mciteSetBstMidEndSepPunct{\mcitedefaultmidpunct}
{\mcitedefaultendpunct}{\mcitedefaultseppunct}\relax
\EndOfBibitem
\bibitem[Ambrose and Walton(1989)Ambrose, and Walton]{Ambrose1989}
Ambrose,~D.; Walton,~J. Vapour pressures up to their critical temperatures of normal alkanes and 1-alkanols. \emph{Pure and Applied Chemistry} \textbf{1989}, \emph{61}, 1395--1403\relax
\mciteBstWouldAddEndPuncttrue
\mciteSetBstMidEndSepPunct{\mcitedefaultmidpunct}
{\mcitedefaultendpunct}{\mcitedefaultseppunct}\relax
\EndOfBibitem
\bibitem[Riedel(1954)]{Riedel1954}
Riedel,~L. Eine neue universelle Dampfdruckformel Untersuchungen über eine Erweiterung des Theorems der übereinstimmenden Zustände. Teil I. \emph{Chemie Ingenieur Technik} \textbf{1954}, \emph{26}, 83--89\relax
\mciteBstWouldAddEndPuncttrue
\mciteSetBstMidEndSepPunct{\mcitedefaultmidpunct}
{\mcitedefaultendpunct}{\mcitedefaultseppunct}\relax
\EndOfBibitem
\bibitem[Macknick and Prausnitz(1979)Macknick, and Prausnitz]{Macknick1979}
Macknick,~A.~B.; Prausnitz,~J.~M. Vapor Pressures of Heavy Liquid Hydrocarbons by a Group-Contribution Method. \emph{Industrial \& Engineering Chemistry Fundamentals} \textbf{1979}, \emph{18}, 348--351\relax
\mciteBstWouldAddEndPuncttrue
\mciteSetBstMidEndSepPunct{\mcitedefaultmidpunct}
{\mcitedefaultendpunct}{\mcitedefaultseppunct}\relax
\EndOfBibitem
\bibitem[Edwards and Prausnitz(1981)Edwards, and Prausnitz]{Edwards1981}
Edwards,~D.~R.; Prausnitz,~J.~M. Estimation of vapor pressures of heavy liquid hydrocarbons containing nitrogen or sulfur by a group-contribution method. \emph{Industrial \& Engineering Chemistry Fundamentals} \textbf{1981}, \emph{20}, 280--283\relax
\mciteBstWouldAddEndPuncttrue
\mciteSetBstMidEndSepPunct{\mcitedefaultmidpunct}
{\mcitedefaultendpunct}{\mcitedefaultseppunct}\relax
\EndOfBibitem
\bibitem[Burkhard(1985)]{Burkhard1985}
Burkhard,~L.~P. Estimation of vapor pressures for halogenated aromatic hydrocarbons by a group-contribution method. \emph{Industrial \& Engineering Chemistry Fundamentals} \textbf{1985}, \emph{24}, 119--120\relax
\mciteBstWouldAddEndPuncttrue
\mciteSetBstMidEndSepPunct{\mcitedefaultmidpunct}
{\mcitedefaultendpunct}{\mcitedefaultseppunct}\relax
\EndOfBibitem
\bibitem[Tu(1994)]{Tu1994}
Tu,~C.-H. Group-contribution method for the estimation of vapor pressures. \emph{Fluid Phase Equilibria} \textbf{1994}, \emph{99}, 105--120\relax
\mciteBstWouldAddEndPuncttrue
\mciteSetBstMidEndSepPunct{\mcitedefaultmidpunct}
{\mcitedefaultendpunct}{\mcitedefaultseppunct}\relax
\EndOfBibitem
\bibitem[Sawaya \latin{et~al.}(2004)Sawaya, Mokbel, Rauzy, Saab, Berro, and Jose]{Sawaya2004}
Sawaya,~T.; Mokbel,~I.; Rauzy,~E.; Saab,~J.; Berro,~C.; Jose,~J. Experimental vapor pressures of alkyl and aryl sulfides - Prediction by a group contribution method. \emph{Fluid Phase Equilibria} \textbf{2004}, \emph{226}, 283--288\relax
\mciteBstWouldAddEndPuncttrue
\mciteSetBstMidEndSepPunct{\mcitedefaultmidpunct}
{\mcitedefaultendpunct}{\mcitedefaultseppunct}\relax
\EndOfBibitem
\bibitem[Asher and Pankow(2006)Asher, and Pankow]{Asher2006}
Asher,~W.~E.; Pankow,~J.~F. Vapor pressure prediction for alkenoic and aromatic organic compounds by a UNIFAC-based group contribution method. \emph{Atmospheric Environment} \textbf{2006}, \emph{40}, 3588--3600\relax
\mciteBstWouldAddEndPuncttrue
\mciteSetBstMidEndSepPunct{\mcitedefaultmidpunct}
{\mcitedefaultendpunct}{\mcitedefaultseppunct}\relax
\EndOfBibitem
\bibitem[Pankow and Asher(2008)Pankow, and Asher]{Pankow2008}
Pankow,~J.~F.; Asher,~W.~E. SIMPOL.1: a simple group contribution method for predicting vapor pressures and enthalpies of vaporization of multifunctional organic compounds. \emph{Atmospheric Chemistry and Physics} \textbf{2008}, \emph{8}, 2773--2796\relax
\mciteBstWouldAddEndPuncttrue
\mciteSetBstMidEndSepPunct{\mcitedefaultmidpunct}
{\mcitedefaultendpunct}{\mcitedefaultseppunct}\relax
\EndOfBibitem
\bibitem[Nannoolal \latin{et~al.}(2008)Nannoolal, Rarey, and Ramjugernath]{Nannoolal2008}
Nannoolal,~Y.; Rarey,~J.; Ramjugernath,~D. Estimation of pure component properties: Part 3. Estimation of the vapor pressure of non-electrolyte organic compounds via group contributions and group interactions. \emph{Fluid Phase Equilibria} \textbf{2008}, \emph{269}, 117--133\relax
\mciteBstWouldAddEndPuncttrue
\mciteSetBstMidEndSepPunct{\mcitedefaultmidpunct}
{\mcitedefaultendpunct}{\mcitedefaultseppunct}\relax
\EndOfBibitem
\bibitem[Moller \latin{et~al.}(2008)Moller, Rarey, and Ramjugernath]{Moller2008}
Moller,~B.; Rarey,~J.; Ramjugernath,~D. Estimation of the vapour pressure of non-electrolyte organic compounds via group contributions and group interactions. \emph{Journal of Molecular Liquids} \textbf{2008}, \emph{143}, 52--63\relax
\mciteBstWouldAddEndPuncttrue
\mciteSetBstMidEndSepPunct{\mcitedefaultmidpunct}
{\mcitedefaultendpunct}{\mcitedefaultseppunct}\relax
\EndOfBibitem
\bibitem[Ceriani \latin{et~al.}(2013)Ceriani, Gani, and Liu]{Ceriani2013}
Ceriani,~R.; Gani,~R.; Liu,~Y. Prediction of vapor pressure and heats of vaporization of edible oil/fat compounds by group contribution. \emph{Fluid Phase Equilibria} \textbf{2013}, \emph{337}, 53--59\relax
\mciteBstWouldAddEndPuncttrue
\mciteSetBstMidEndSepPunct{\mcitedefaultmidpunct}
{\mcitedefaultendpunct}{\mcitedefaultseppunct}\relax
\EndOfBibitem
\bibitem[Rezakazemi \latin{et~al.}(2013)Rezakazemi, Marjani, and Shirazian]{Rezakazemi2013}
Rezakazemi,~M.; Marjani,~A.; Shirazian,~S. Development of a Group Contribution Method Based on UNIFAC Groups for the Estimation of Vapor Pressures of Pure Hydrocarbon Compounds. \emph{Chemical Engineering \& Technology} \textbf{2013}, \emph{36}, 483--491\relax
\mciteBstWouldAddEndPuncttrue
\mciteSetBstMidEndSepPunct{\mcitedefaultmidpunct}
{\mcitedefaultendpunct}{\mcitedefaultseppunct}\relax
\EndOfBibitem
\bibitem[Wang \latin{et~al.}(2015)Wang, Meng, Jia, and Song]{Wang2015a}
Wang,~T.-Y.; Meng,~X.-Z.; Jia,~M.; Song,~X.-C. Predicting the vapor pressure of fatty acid esters in biodiesel by group contribution method. \emph{Fuel Processing Technology} \textbf{2015}, \emph{131}, 223--229\relax
\mciteBstWouldAddEndPuncttrue
\mciteSetBstMidEndSepPunct{\mcitedefaultmidpunct}
{\mcitedefaultendpunct}{\mcitedefaultseppunct}\relax
\EndOfBibitem
\bibitem[Beck \latin{et~al.}(2000)Beck, Breindl, and Clark]{Beck2000}
Beck,~B.; Breindl,~A.; Clark,~T. QM/NN QSPR Models with Error Estimation: Vapor Pressure and LogP. \emph{Journal of Chemical Information and Computer Sciences} \textbf{2000}, \emph{40}, 1046--1051\relax
\mciteBstWouldAddEndPuncttrue
\mciteSetBstMidEndSepPunct{\mcitedefaultmidpunct}
{\mcitedefaultendpunct}{\mcitedefaultseppunct}\relax
\EndOfBibitem
\bibitem[Gharagheizi \latin{et~al.}(2012)Gharagheizi, Eslamimanesh, Ilani-Kashkouli, Mohammadi, and Richon]{Gharagheizi2012}
Gharagheizi,~F.; Eslamimanesh,~A.; Ilani-Kashkouli,~P.; Mohammadi,~A.~H.; Richon,~D. QSPR molecular approach for representation/prediction of very large vapor pressure dataset. \emph{Chemical Engineering Science} \textbf{2012}, \emph{76}, 99--107\relax
\mciteBstWouldAddEndPuncttrue
\mciteSetBstMidEndSepPunct{\mcitedefaultmidpunct}
{\mcitedefaultendpunct}{\mcitedefaultseppunct}\relax
\EndOfBibitem
\bibitem[Katritzky \latin{et~al.}(2007)Katritzky, Slavov, Dobchev, and Karelson]{Katritzky2007}
Katritzky,~A.~R.; Slavov,~S.~H.; Dobchev,~D.~A.; Karelson,~M. Rapid QSPR model development technique for prediction of vapor pressure of organic compounds. \emph{Computers \& Chemical Engineering} \textbf{2007}, \emph{31}, 1123--1130\relax
\mciteBstWouldAddEndPuncttrue
\mciteSetBstMidEndSepPunct{\mcitedefaultmidpunct}
{\mcitedefaultendpunct}{\mcitedefaultseppunct}\relax
\EndOfBibitem
\bibitem[Katritzky \latin{et~al.}(1995)Katritzky, Lobanov, and Karelson]{Katritzky1995}
Katritzky,~A.~R.; Lobanov,~V.~S.; Karelson,~M. QSPR: the correlation and quantitative prediction of chemical and physical properties from structure. \emph{Chemical Society Reviews} \textbf{1995}, \emph{24}, 279\relax
\mciteBstWouldAddEndPuncttrue
\mciteSetBstMidEndSepPunct{\mcitedefaultmidpunct}
{\mcitedefaultendpunct}{\mcitedefaultseppunct}\relax
\EndOfBibitem
\bibitem[Hsieh and Lin(2008)Hsieh, and Lin]{Hsieh2008}
Hsieh,~C.; Lin,~S. Determination of cubic equation of state parameters for pure fluids from first principle solvation calculations. \emph{AIChE Journal} \textbf{2008}, \emph{54}, 2174--2181\relax
\mciteBstWouldAddEndPuncttrue
\mciteSetBstMidEndSepPunct{\mcitedefaultmidpunct}
{\mcitedefaultendpunct}{\mcitedefaultseppunct}\relax
\EndOfBibitem
\bibitem[Wang \latin{et~al.}(2015)Wang, Hsieh, and Lin]{Wang2015}
Wang,~L.-H.; Hsieh,~C.-M.; Lin,~S.-T. Improved Prediction of Vapor Pressure for Pure Liquids and Solids from the PR+COSMOSAC Equation of State. \emph{Industrial \& Engineering Chemistry Research} \textbf{2015}, \emph{54}, 10115--10125\relax
\mciteBstWouldAddEndPuncttrue
\mciteSetBstMidEndSepPunct{\mcitedefaultmidpunct}
{\mcitedefaultendpunct}{\mcitedefaultseppunct}\relax
\EndOfBibitem
\bibitem[Liang \latin{et~al.}(2019)Liang, Li, Wang, Lin, and Hsieh]{Liang2019}
Liang,~H.-H.; Li,~J.-Y.; Wang,~L.-H.; Lin,~S.-T.; Hsieh,~C.-M. Improvement to PR+COSMOSAC EOS for Predicting the Vapor Pressure of Nonelectrolyte Organic Solids and Liquids. \emph{Industrial \& Engineering Chemistry Research} \textbf{2019}, \emph{58}, 5030--5040\relax
\mciteBstWouldAddEndPuncttrue
\mciteSetBstMidEndSepPunct{\mcitedefaultmidpunct}
{\mcitedefaultendpunct}{\mcitedefaultseppunct}\relax
\EndOfBibitem
\bibitem[Habicht \latin{et~al.}(2023)Habicht, Brandenbusch, and Sadowski]{Habicht2023}
Habicht,~J.; Brandenbusch,~C.; Sadowski,~G. Predicting PC-SAFT pure-component parameters by machine learning using a molecular fingerprint as key input. \emph{Fluid Phase Equilibria} \textbf{2023}, \emph{565}, 113657\relax
\mciteBstWouldAddEndPuncttrue
\mciteSetBstMidEndSepPunct{\mcitedefaultmidpunct}
{\mcitedefaultendpunct}{\mcitedefaultseppunct}\relax
\EndOfBibitem
\bibitem[Felton \latin{et~al.}(2024)Felton, Raßpe-Lange, Rittig, Leonhard, Mitsos, Meyer-Kirschner, Knösche, and Lapkin]{Felton2024}
Felton,~K.~C.; Raßpe-Lange,~L.; Rittig,~J.~G.; Leonhard,~K.; Mitsos,~A.; Meyer-Kirschner,~J.; Knösche,~C.; Lapkin,~A.~A. ML-SAFT: A machine learning framework for PCP-SAFT parameter prediction. \emph{Chemical Engineering Journal} \textbf{2024}, \emph{492}, 151999\relax
\mciteBstWouldAddEndPuncttrue
\mciteSetBstMidEndSepPunct{\mcitedefaultmidpunct}
{\mcitedefaultendpunct}{\mcitedefaultseppunct}\relax
\EndOfBibitem
\bibitem[Winter \latin{et~al.}(2023)Winter, Rehner, Esper, Schilling, and Bardow]{Winter2023}
Winter,~B.; Rehner,~P.; Esper,~T.; Schilling,~J.; Bardow,~A. Understanding the language of molecules: Predicting pure component parameters for the PC-SAFT equation of state from SMILES. arXiv preprint, 2023; 2309.12404\relax
\mciteBstWouldAddEndPuncttrue
\mciteSetBstMidEndSepPunct{\mcitedefaultmidpunct}
{\mcitedefaultendpunct}{\mcitedefaultseppunct}\relax
\EndOfBibitem
\bibitem[Klamt and Eckert(2000)Klamt, and Eckert]{Klamt2000}
Klamt,~A.; Eckert,~F. COSMO-RS: a novel and efficient method for the a priori prediction of thermophysical data of liquids. \emph{Fluid Phase Equilibria} \textbf{2000}, \emph{172}, 43--72\relax
\mciteBstWouldAddEndPuncttrue
\mciteSetBstMidEndSepPunct{\mcitedefaultmidpunct}
{\mcitedefaultendpunct}{\mcitedefaultseppunct}\relax
\EndOfBibitem
\bibitem[Bell \latin{et~al.}(2020)Bell, Mickoleit, Hsieh, Lin, Vrabec, Breitkopf, and Jäger]{Bell2020}
Bell,~I.~H.; Mickoleit,~E.; Hsieh,~C.-M.; Lin,~S.-T.; Vrabec,~J.; Breitkopf,~C.; Jäger,~A. A Benchmark Open-Source Implementation of COSMO-SAC. \emph{Journal of Chemical Theory and Computation} \textbf{2020}, \emph{16}, 2635--2646\relax
\mciteBstWouldAddEndPuncttrue
\mciteSetBstMidEndSepPunct{\mcitedefaultmidpunct}
{\mcitedefaultendpunct}{\mcitedefaultseppunct}\relax
\EndOfBibitem
\bibitem[Santana \latin{et~al.}(2024)Santana, Rebello, Queiroz, Ribeiro, Shardt, and Nogueira]{Santana2024}
Santana,~V.~V.; Rebello,~C.~M.; Queiroz,~L.~P.; Ribeiro,~A.~M.; Shardt,~N.; Nogueira,~I.~B. PUFFIN: A path-unifying feed-forward interfaced network for vapor pressure prediction. \emph{Chemical Engineering Science} \textbf{2024}, \emph{286}, 119623\relax
\mciteBstWouldAddEndPuncttrue
\mciteSetBstMidEndSepPunct{\mcitedefaultmidpunct}
{\mcitedefaultendpunct}{\mcitedefaultseppunct}\relax
\EndOfBibitem
\bibitem[Lin \latin{et~al.}(2024)Lin, Liang, Lin, and Li]{Lin2024}
Lin,~Y.-H.; Liang,~H.-H.; Lin,~S.-T.; Li,~Y.-P. Advancing vapor pressure prediction: A machine learning approach with directed message passing neural networks. \emph{Journal of the Taiwan Institute of Chemical Engineers} \textbf{2024}, 105926\relax
\mciteBstWouldAddEndPuncttrue
\mciteSetBstMidEndSepPunct{\mcitedefaultmidpunct}
{\mcitedefaultendpunct}{\mcitedefaultseppunct}\relax
\EndOfBibitem
\bibitem[Sanchez~Medina \latin{et~al.}(2022)Sanchez~Medina, Linke, Stoll, and Sundmacher]{SanchezMedina2022}
Sanchez~Medina,~E.~I.; Linke,~S.; Stoll,~M.; Sundmacher,~K. Graph neural networks for the prediction of infinite dilution activity coefficients. \emph{Digital Discovery} \textbf{2022}, \emph{1}, 216--225\relax
\mciteBstWouldAddEndPuncttrue
\mciteSetBstMidEndSepPunct{\mcitedefaultmidpunct}
{\mcitedefaultendpunct}{\mcitedefaultseppunct}\relax
\EndOfBibitem
\bibitem[Sanchez~Medina \latin{et~al.}(2023)Sanchez~Medina, Linke, Stoll, and Sundmacher]{SanchezMedina2023}
Sanchez~Medina,~E.~I.; Linke,~S.; Stoll,~M.; Sundmacher,~K. Gibbs–Helmholtz graph neural network: capturing the temperature dependency of activity coefficients at infinite dilution. \emph{Digital Discovery} \textbf{2023}, \emph{2}, 781--798\relax
\mciteBstWouldAddEndPuncttrue
\mciteSetBstMidEndSepPunct{\mcitedefaultmidpunct}
{\mcitedefaultendpunct}{\mcitedefaultseppunct}\relax
\EndOfBibitem
\bibitem[Ahmad \latin{et~al.}(2023)Ahmad, Tayara, and Chong]{Ahmad2023}
Ahmad,~W.; Tayara,~H.; Chong,~K.~T. Attention-Based Graph Neural Network for Molecular Solubility Prediction. \emph{ACS Omega} \textbf{2023}, \emph{8}, 3236--3244\relax
\mciteBstWouldAddEndPuncttrue
\mciteSetBstMidEndSepPunct{\mcitedefaultmidpunct}
{\mcitedefaultendpunct}{\mcitedefaultseppunct}\relax
\EndOfBibitem
\bibitem[Qu \latin{et~al.}(2022)Qu, Kearsley, Schneider, Keyrouz, and Allison]{Qu2022}
Qu,~C.; Kearsley,~A.~J.; Schneider,~B.~I.; Keyrouz,~W.; Allison,~T.~C. Graph convolutional neural network applied to the prediction of normal boiling point. \emph{Journal of Molecular Graphics and Modelling} \textbf{2022}, \emph{112}, 108149\relax
\mciteBstWouldAddEndPuncttrue
\mciteSetBstMidEndSepPunct{\mcitedefaultmidpunct}
{\mcitedefaultendpunct}{\mcitedefaultseppunct}\relax
\EndOfBibitem
\bibitem[Aouichaoui \latin{et~al.}(2023)Aouichaoui, Cogliati, Abildskov, and Sin]{Aouichaoui2023}
Aouichaoui,~A.~R.; Cogliati,~A.; Abildskov,~J.; Sin,~G. \emph{33rd European Symposium on Computer Aided Process Engineering}; Elsevier, 2023; pp 575--581\relax
\mciteBstWouldAddEndPuncttrue
\mciteSetBstMidEndSepPunct{\mcitedefaultmidpunct}
{\mcitedefaultendpunct}{\mcitedefaultseppunct}\relax
\EndOfBibitem
\bibitem[Aouichaoui \latin{et~al.}(2023)Aouichaoui, Fan, Mansouri, Abildskov, and Sin]{Aouichaoui2023a}
Aouichaoui,~A. R.~N.; Fan,~F.; Mansouri,~S.~S.; Abildskov,~J.; Sin,~G. Combining Group-Contribution Concept and Graph Neural Networks Toward Interpretable Molecular Property Models. \emph{Journal of Chemical Information and Modeling} \textbf{2023}, \emph{63}, 725--744\relax
\mciteBstWouldAddEndPuncttrue
\mciteSetBstMidEndSepPunct{\mcitedefaultmidpunct}
{\mcitedefaultendpunct}{\mcitedefaultseppunct}\relax
\EndOfBibitem
\bibitem[Aouichaoui \latin{et~al.}(2023)Aouichaoui, Fan, Abildskov, and Sin]{Aouichaoui2023b}
Aouichaoui,~A.~R.; Fan,~F.; Abildskov,~J.; Sin,~G. Application of interpretable group-embedded graph neural networks for pure compound properties. \emph{Computers \& Chemical Engineering} \textbf{2023}, \emph{176}, 108291\relax
\mciteBstWouldAddEndPuncttrue
\mciteSetBstMidEndSepPunct{\mcitedefaultmidpunct}
{\mcitedefaultendpunct}{\mcitedefaultseppunct}\relax
\EndOfBibitem
\bibitem[Hayer \latin{et~al.}(2024)Hayer, Hasse, and Jirasek]{Hayer2024a}
Hayer,~N.; Hasse,~H.; Jirasek,~F. Prediction of Temperature-Dependent Henry’s Law Constants by Matrix Completion. \emph{The Journal of Physical Chemistry B} \textbf{2024}, \emph{129}, 409--416\relax
\mciteBstWouldAddEndPuncttrue
\mciteSetBstMidEndSepPunct{\mcitedefaultmidpunct}
{\mcitedefaultendpunct}{\mcitedefaultseppunct}\relax
\EndOfBibitem
\bibitem[Hayer \latin{et~al.}(2025)Hayer, Wendel, Mandt, Hasse, and Jirasek]{Hayer2025}
Hayer,~N.; Wendel,~T.; Mandt,~S.; Hasse,~H.; Jirasek,~F. Advancing thermodynamic group-contribution methods by machine learning: UNIFAC 2.0. \emph{Chemical Engineering Journal} \textbf{2025}, \emph{504}, 158667\relax
\mciteBstWouldAddEndPuncttrue
\mciteSetBstMidEndSepPunct{\mcitedefaultmidpunct}
{\mcitedefaultendpunct}{\mcitedefaultseppunct}\relax
\EndOfBibitem
\bibitem[Specht \latin{et~al.}(2024)Specht, Nagda, Fellenz, Mandt, Hasse, and Jirasek]{Specht2024}
Specht,~T.; Nagda,~M.; Fellenz,~S.; Mandt,~S.; Hasse,~H.; Jirasek,~F. HANNA: hard-constraint neural network for consistent activity coefficient prediction. \emph{Chemical Science} \textbf{2024}, \emph{15}, 19777--19786\relax
\mciteBstWouldAddEndPuncttrue
\mciteSetBstMidEndSepPunct{\mcitedefaultmidpunct}
{\mcitedefaultendpunct}{\mcitedefaultseppunct}\relax
\EndOfBibitem
\bibitem[Damay \latin{et~al.}(2021)Damay, Jirasek, Kloft, Bortz, and Hasse]{Damay2021}
Damay,~J.; Jirasek,~F.; Kloft,~M.; Bortz,~M.; Hasse,~H. Predicting Activity Coefficients at Infinite Dilution for Varying Temperatures by Matrix Completion. \emph{Industrial \& Engineering Chemistry Research} \textbf{2021}, \emph{60}, 14564--14578\relax
\mciteBstWouldAddEndPuncttrue
\mciteSetBstMidEndSepPunct{\mcitedefaultmidpunct}
{\mcitedefaultendpunct}{\mcitedefaultseppunct}\relax
\EndOfBibitem
\bibitem[Rittig \latin{et~al.}(2023)Rittig, Felton, Lapkin, and Mitsos]{Rittig2023}
Rittig,~J.~G.; Felton,~K.~C.; Lapkin,~A.~A.; Mitsos,~A. Gibbs-Duhem-Informed Neural Networks for Binary Activity Coefficient Prediction. \emph{Digital Discovery} \textbf{2023}, \emph{2}, 1752--1767\relax
\mciteBstWouldAddEndPuncttrue
\mciteSetBstMidEndSepPunct{\mcitedefaultmidpunct}
{\mcitedefaultendpunct}{\mcitedefaultseppunct}\relax
\EndOfBibitem
\bibitem[Rittig \latin{et~al.}(2023)Rittig, Ben~Hicham, Schweidtmann, Dahmen, and Mitsos]{Rittig2023a}
Rittig,~J.~G.; Ben~Hicham,~K.; Schweidtmann,~A.~M.; Dahmen,~M.; Mitsos,~A. Graph neural networks for temperature-dependent activity coefficient prediction of solutes in ionic liquids. \emph{Computers \& Chemical Engineering} \textbf{2023}, \emph{171}, 108153\relax
\mciteBstWouldAddEndPuncttrue
\mciteSetBstMidEndSepPunct{\mcitedefaultmidpunct}
{\mcitedefaultendpunct}{\mcitedefaultseppunct}\relax
\EndOfBibitem
\bibitem[Jirasek and Hasse(2023)Jirasek, and Hasse]{Jirasek2023}
Jirasek,~F.; Hasse,~H. Combining Machine Learning with Physical Knowledge in Thermodynamic Modeling of Fluid Mixtures. \emph{Annual Review of Chemical and Biomolecular Engineering} \textbf{2023}, \emph{14}, 31--51\relax
\mciteBstWouldAddEndPuncttrue
\mciteSetBstMidEndSepPunct{\mcitedefaultmidpunct}
{\mcitedefaultendpunct}{\mcitedefaultseppunct}\relax
\EndOfBibitem
\bibitem[Rogers and Hahn(2010)Rogers, and Hahn]{Rogers2010}
Rogers,~D.; Hahn,~M. Extended-Connectivity Fingerprints. \emph{Journal of Chemical Information and Modeling} \textbf{2010}, \emph{50}, 742--754\relax
\mciteBstWouldAddEndPuncttrue
\mciteSetBstMidEndSepPunct{\mcitedefaultmidpunct}
{\mcitedefaultendpunct}{\mcitedefaultseppunct}\relax
\EndOfBibitem
\bibitem[Gross and Sadowski(2001)Gross, and Sadowski]{Gross2001}
Gross,~J.; Sadowski,~G. Perturbed-Chain SAFT: An Equation of State Based on a Perturbation Theory for Chain Molecules. \emph{Industrial \& Engineering Chemistry Research} \textbf{2001}, \emph{40}, 1244--1260\relax
\mciteBstWouldAddEndPuncttrue
\mciteSetBstMidEndSepPunct{\mcitedefaultmidpunct}
{\mcitedefaultendpunct}{\mcitedefaultseppunct}\relax
\EndOfBibitem
\bibitem[Gross and Vrabec(2005)Gross, and Vrabec]{Gross2005}
Gross,~J.; Vrabec,~J. An equation‐of‐state contribution for polar components: Dipolar molecules. \emph{AIChE Journal} \textbf{2005}, \emph{52}, 1194--1204\relax
\mciteBstWouldAddEndPuncttrue
\mciteSetBstMidEndSepPunct{\mcitedefaultmidpunct}
{\mcitedefaultendpunct}{\mcitedefaultseppunct}\relax
\EndOfBibitem
\bibitem[Weininger(1988)]{Weininger1988}
Weininger,~D. SMILES, a chemical language and information system. 1. Introduction to methodology and encoding rules. \emph{Journal of Chemical Information and Computer Sciences} \textbf{1988}, \emph{28}, 31--36\relax
\mciteBstWouldAddEndPuncttrue
\mciteSetBstMidEndSepPunct{\mcitedefaultmidpunct}
{\mcitedefaultendpunct}{\mcitedefaultseppunct}\relax
\EndOfBibitem
\bibitem[Duvenaud \latin{et~al.}(2015)Duvenaud, Maclaurin, Aguilera-Iparraguirre, Gómez-Bombarelli, Hirzel, Aspuru-Guzik, and Adams]{Duvenaud2015}
Duvenaud,~D.; Maclaurin,~D.; Aguilera-Iparraguirre,~J.; Gómez-Bombarelli,~R.; Hirzel,~T.; Aspuru-Guzik,~A.; Adams,~R.~P. Convolutional Networks on Graphs for Learning Molecular Fingerprints. arXiv preprint, 2015; 1509.09292\relax
\mciteBstWouldAddEndPuncttrue
\mciteSetBstMidEndSepPunct{\mcitedefaultmidpunct}
{\mcitedefaultendpunct}{\mcitedefaultseppunct}\relax
\EndOfBibitem
\bibitem[Yang \latin{et~al.}(2019)Yang, Swanson, Jin, Coley, Eiden, Gao, Guzman-Perez, Hopper, Kelley, Mathea, Palmer, Settels, Jaakkola, Jensen, and Barzilay]{Yang2019}
Yang,~K.; Swanson,~K.; Jin,~W.; Coley,~C.; Eiden,~P.; Gao,~H.; Guzman-Perez,~A.; Hopper,~T.; Kelley,~B.; Mathea,~M.; Palmer,~A.; Settels,~V.; Jaakkola,~T.; Jensen,~K.; Barzilay,~R. Analyzing Learned Molecular Representations for Property Prediction. \emph{Journal of Chemical Information and Modeling} \textbf{2019}, \emph{59}, 3370--3388\relax
\mciteBstWouldAddEndPuncttrue
\mciteSetBstMidEndSepPunct{\mcitedefaultmidpunct}
{\mcitedefaultendpunct}{\mcitedefaultseppunct}\relax
\EndOfBibitem
\bibitem[Wang and Hu(2023)Wang, and Hu]{Wang2023}
Wang,~G.; Hu,~P. Prediction of normal boiling point and critical temperature of refrigerants by graph neural network and transfer learning. \emph{International Journal of Refrigeration} \textbf{2023}, \emph{151}, 97--104\relax
\mciteBstWouldAddEndPuncttrue
\mciteSetBstMidEndSepPunct{\mcitedefaultmidpunct}
{\mcitedefaultendpunct}{\mcitedefaultseppunct}\relax
\EndOfBibitem
\bibitem[Veličković \latin{et~al.}(2018)Veličković, Cucurull, Casanova, Romero, Liò, and Bengio]{Velickovic2018}
Veličković,~P.; Cucurull,~G.; Casanova,~A.; Romero,~A.; Liò,~P.; Bengio,~Y. Graph Attention Networks. arXiv preprint, 2018; 1710.10903\relax
\mciteBstWouldAddEndPuncttrue
\mciteSetBstMidEndSepPunct{\mcitedefaultmidpunct}
{\mcitedefaultendpunct}{\mcitedefaultseppunct}\relax
\EndOfBibitem
\bibitem[RDKit, online()]{rdkit}
{RDK}it: Open-source cheminformatics. \url{http://www.rdkit.org}, Version: 2023.03.1\relax
\mciteBstWouldAddEndPuncttrue
\mciteSetBstMidEndSepPunct{\mcitedefaultmidpunct}
{\mcitedefaultendpunct}{\mcitedefaultseppunct}\relax
\EndOfBibitem
\bibitem[Brody \latin{et~al.}(2022)Brody, Alon, and Yahav]{Brody2022}
Brody,~S.; Alon,~U.; Yahav,~E. How Attentive are Graph Attention Networks? International Conference on Learning Representations. 2022\relax
\mciteBstWouldAddEndPuncttrue
\mciteSetBstMidEndSepPunct{\mcitedefaultmidpunct}
{\mcitedefaultendpunct}{\mcitedefaultseppunct}\relax
\EndOfBibitem
\bibitem[Fey and Lenssen(2019)Fey, and Lenssen]{pytorch_geometric}
Fey,~M.; Lenssen,~J.~E. {Fast Graph Representation Learning with PyTorch Geometric}. 2019; \url{https://github.com/pyg-team/pytorch_geometric}\relax
\mciteBstWouldAddEndPuncttrue
\mciteSetBstMidEndSepPunct{\mcitedefaultmidpunct}
{\mcitedefaultendpunct}{\mcitedefaultseppunct}\relax
\EndOfBibitem
\bibitem[Buterez \latin{et~al.}(2023)Buterez, Janet, Kiddle, Oglic, and Liò]{Buterez2023}
Buterez,~D.; Janet,~J.~P.; Kiddle,~S.~J.; Oglic,~D.; Liò,~P. Modelling local and general quantum mechanical properties with attention-based pooling. \emph{Communications Chemistry} \textbf{2023}, \emph{6}\relax
\mciteBstWouldAddEndPuncttrue
\mciteSetBstMidEndSepPunct{\mcitedefaultmidpunct}
{\mcitedefaultendpunct}{\mcitedefaultseppunct}\relax
\EndOfBibitem
\bibitem[Baek \latin{et~al.}(2021)Baek, Kang, and Hwang]{Baek2021}
Baek,~J.; Kang,~M.; Hwang,~S.~J. Accurate Learning of Graph Representations with Graph Multiset Pooling. arXiv preprint, 2021; 2102.11533\relax
\mciteBstWouldAddEndPuncttrue
\mciteSetBstMidEndSepPunct{\mcitedefaultmidpunct}
{\mcitedefaultendpunct}{\mcitedefaultseppunct}\relax
\EndOfBibitem
\bibitem[Vaswani \latin{et~al.}(2017)Vaswani, Shazeer, Parmar, Uszkoreit, Jones, Gomez, Kaiser, and Polosukhin]{Vaswani2017}
Vaswani,~A.; Shazeer,~N.; Parmar,~N.; Uszkoreit,~J.; Jones,~L.; Gomez,~A.~N.; Kaiser,~L.; Polosukhin,~I. Attention Is All You Need. arXiv preprint, 2017; 1706.03762\relax
\mciteBstWouldAddEndPuncttrue
\mciteSetBstMidEndSepPunct{\mcitedefaultmidpunct}
{\mcitedefaultendpunct}{\mcitedefaultseppunct}\relax
\EndOfBibitem
\bibitem[Paszke \latin{et~al.}(2019)Paszke, Gross, Massa, Lerer, Bradbury, Chanan, Killeen, Lin, Gimelshein, Antiga, Desmaison, Köpf, Yang, DeVito, Raison, Tejani, Chilamkurthy, Steiner, Fang, Bai, and Chintala]{pytorch}
Paszke,~A. \latin{et~al.}  PyTorch: An Imperative Style, High-Performance Deep Learning Library. arXiv preprint, 2019; 1912.01703\relax
\mciteBstWouldAddEndPuncttrue
\mciteSetBstMidEndSepPunct{\mcitedefaultmidpunct}
{\mcitedefaultendpunct}{\mcitedefaultseppunct}\relax
\EndOfBibitem
\bibitem[Loshchilov and Hutter(2017)Loshchilov, and Hutter]{Loshchilov2017}
Loshchilov,~I.; Hutter,~F. Decoupled Weight Decay Regularization. arXiv preprint, 2017; 1711.05101\relax
\mciteBstWouldAddEndPuncttrue
\mciteSetBstMidEndSepPunct{\mcitedefaultmidpunct}
{\mcitedefaultendpunct}{\mcitedefaultseppunct}\relax
\EndOfBibitem
\bibitem[Smith and Topin(2017)Smith, and Topin]{Smith2017}
Smith,~L.~N.; Topin,~N. Super-Convergence: Very Fast Training of Neural Networks Using Large Learning Rates. arXiv preprint, 2017; 1708.07120\relax
\mciteBstWouldAddEndPuncttrue
\mciteSetBstMidEndSepPunct{\mcitedefaultmidpunct}
{\mcitedefaultendpunct}{\mcitedefaultseppunct}\relax
\EndOfBibitem
\bibitem[Nannoolal \latin{et~al.}(2004)Nannoolal, Rarey, Ramjugernath, and Cordes]{Nannoolal2004}
Nannoolal,~Y.; Rarey,~J.; Ramjugernath,~D.; Cordes,~W. Estimation of pure component properties: Part 1. Estimation of the normal boiling point of non-electrolyte organic compounds via group contributions and group interactions. \emph{Fluid Phase Equilibria} \textbf{2004}, \emph{226}, 45--63\relax
\mciteBstWouldAddEndPuncttrue
\mciteSetBstMidEndSepPunct{\mcitedefaultmidpunct}
{\mcitedefaultendpunct}{\mcitedefaultseppunct}\relax
\EndOfBibitem
\end{mcitethebibliography}
\end{document}


\section{Details on Model Optimization}
Tab.~\ref{tab:hyperparameters_gridsearch} lists the hyperparameters of the GRAPPA architecture and the training process. The hyperparameters for which sets are specified were systematically varied during the grid search, and the bold values indicate the final choices selected based on the validation scores. The remaining parameters were not varied but set as specified. 
\begin{table}[ht]
\centering
\caption{Overview of the hyperparameters of GRAPPA. Hyperparameters for which sets are specified were varied during the grid search, whereby the bold values indicate the final choices selected based on the validation scores.}
\label{tab:hyperparameters_gridsearch}
\begin{tabular}{ll}
\hline
\textbf{Hyperparameter}      & \textbf{Values}            \\
\hline
Number of GNN layers                   & \{2, 3, \textbf{4}, 5\}                      \\
Number of GNN heads                    & \{1, \textbf{2}, 3, 4, 5\}                   \\
GNN convolutional dimension  & 32                               \\
Number of hidden layers                & \{1, 2, \textbf{3}\}                         \\
Neurons in the hidden layers        & 16                                \\
Pooling method               & \{\textit{sum}, \textbf{\textit{interaction}}\}\\

OneCycleLR \texttt{max\_lr}   &  0.001                      \\
ReduceLROnPlateau \texttt{factor} & 0.5                    \\
ReduceLROnPlateau \texttt{patience} & 5                \\
Batch size                   & 512                                   \\
Huber loss \texttt{delta}   & 0.5 \\
Number of epochs in warm-up training                      & 100    \\
Number of epochs in main training                       & 100  \\
\hline
\end{tabular}
\end{table}

\section{Detailed Results of the Vapor Pressure Prediction}
\begin{figure}
    \begin{subfigure}{0.49\textwidth}
        \includegraphics[width=\linewidth]{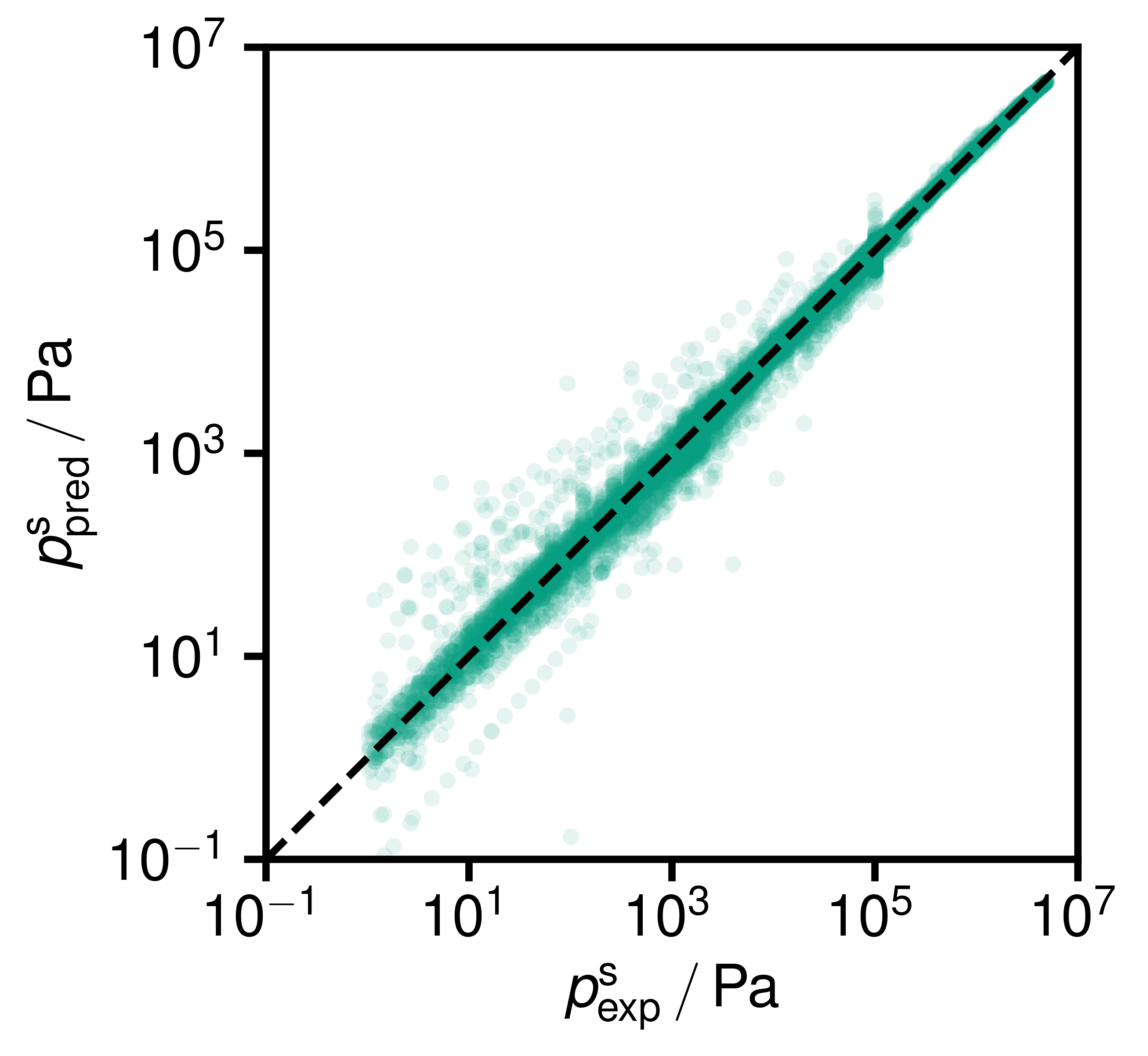}
    \end{subfigure}
    \hfill
    \begin{subfigure}{0.49\textwidth}
        \includegraphics[width=\linewidth]{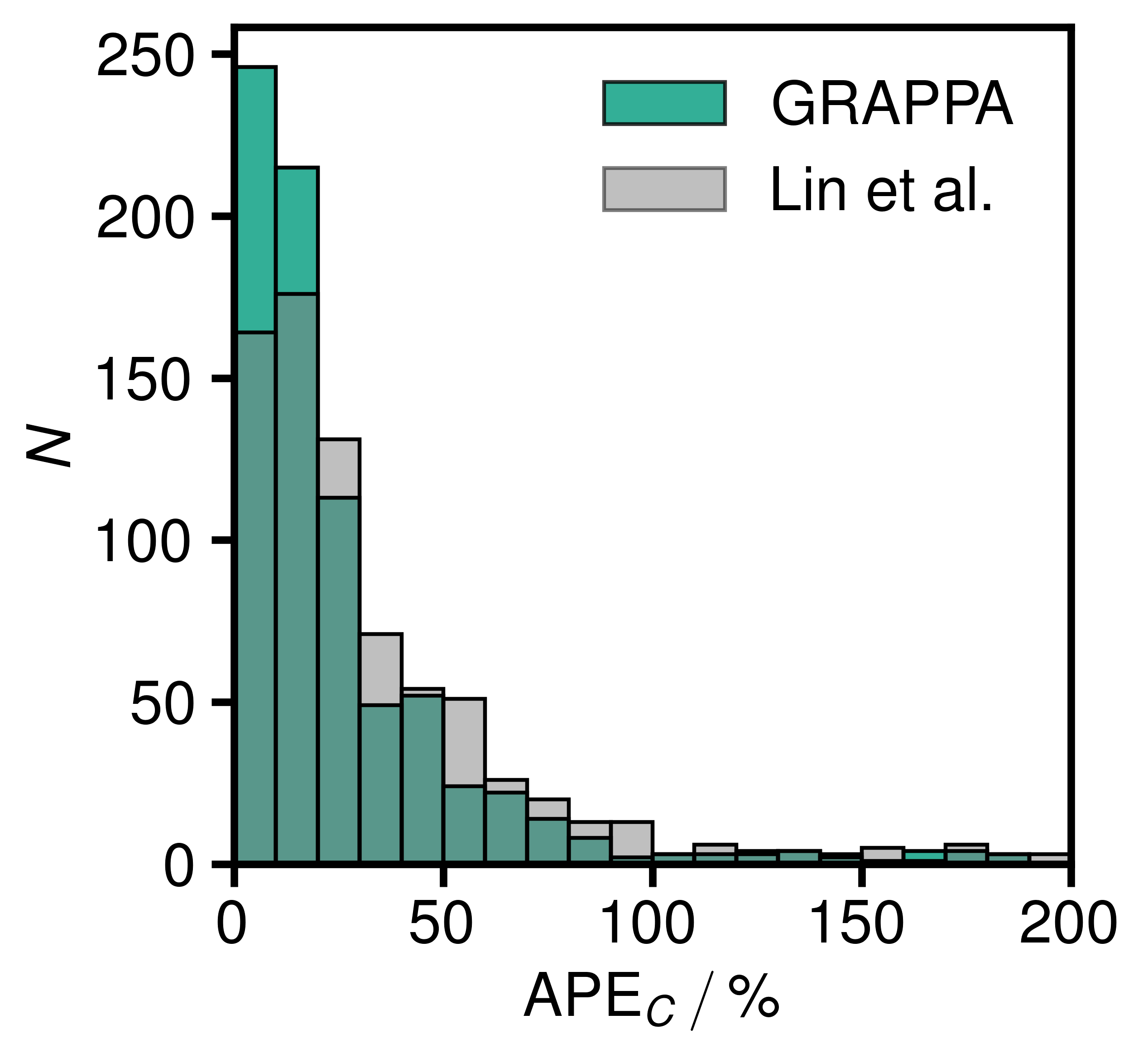}
    \end{subfigure}
    \caption{Left: Parity plot showing the predicted vapor pressure for our test set over the experimental values for GRAPPA. The dashed line marks perfect predictions. Right: Histogram showing the number of components over the $\mathrm{APE}_C$ for GRAPPA and the method by Lin et al.\cite{Lin2024} Only results for components with at least two data points in the test set are shown in both plots. The interval displayed in the histogram covers $96.3\,\%$ and $94.5\,\%$ of the considered components for GRAPPA and the method by Lin et al., respectively.}
    \label{fig:parity_plot_histogram_GRAPPA_Lin2024}
\end{figure}

\begin{figure}
    \begin{subfigure}{0.49\textwidth}
        \centering
        \includegraphics[width=\linewidth]{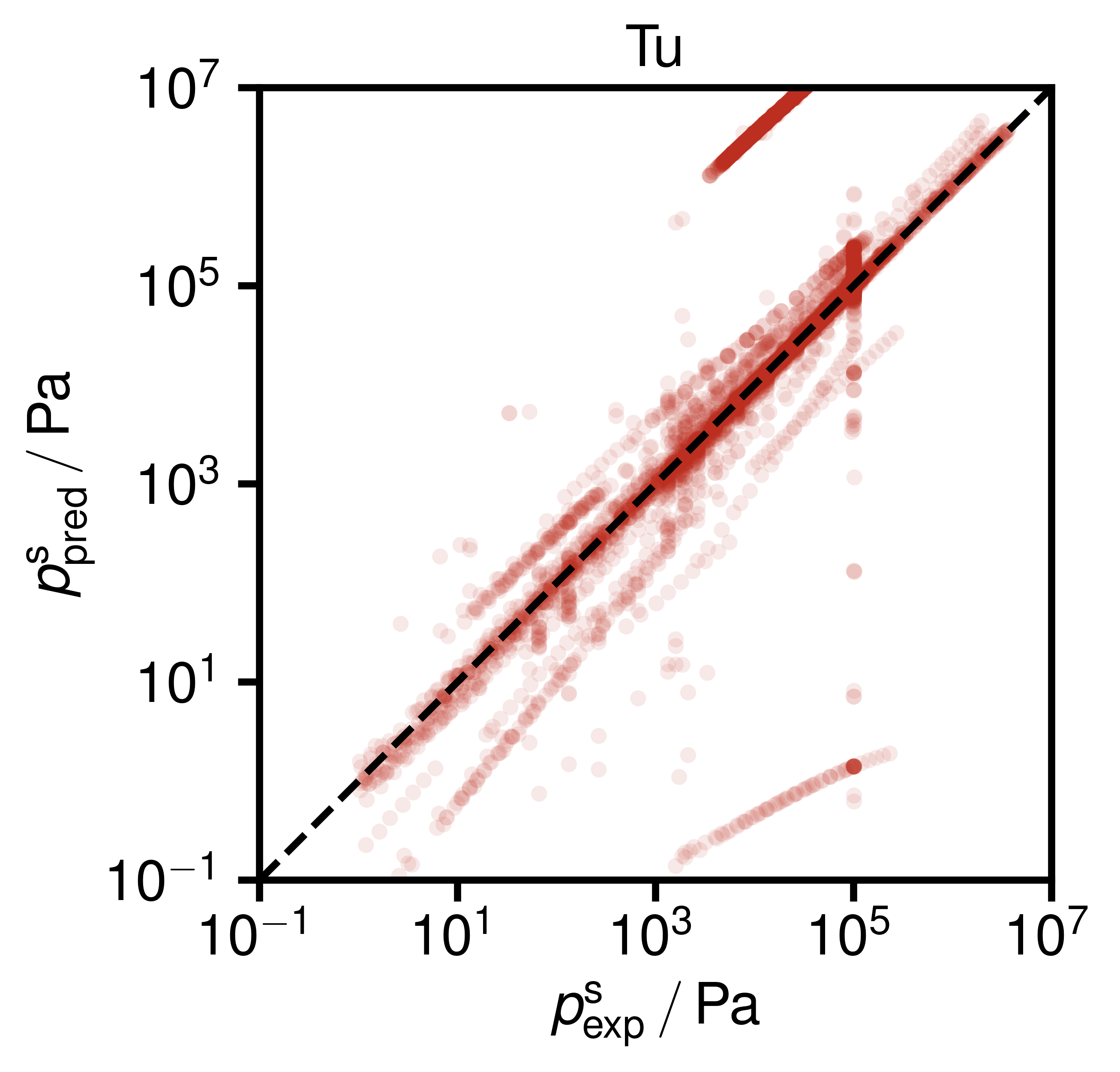}
    \end{subfigure}
    \hfill
    \begin{subfigure}{0.49\textwidth}
        \centering
        \includegraphics[width=\linewidth]{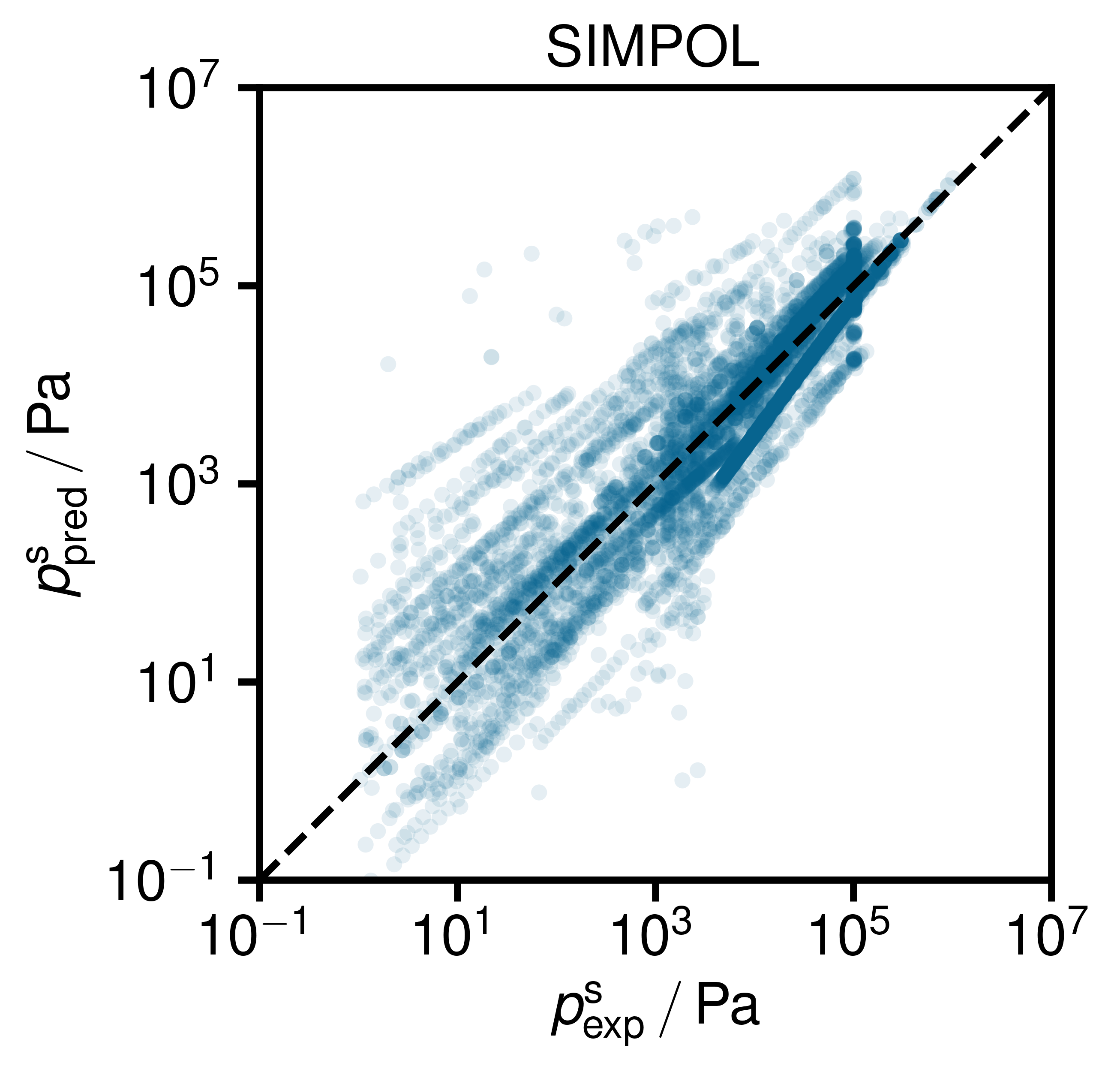}
    \end{subfigure}
    \\
    \begin{subfigure}{0.49\textwidth}
        \centering
        \includegraphics[width=\linewidth]{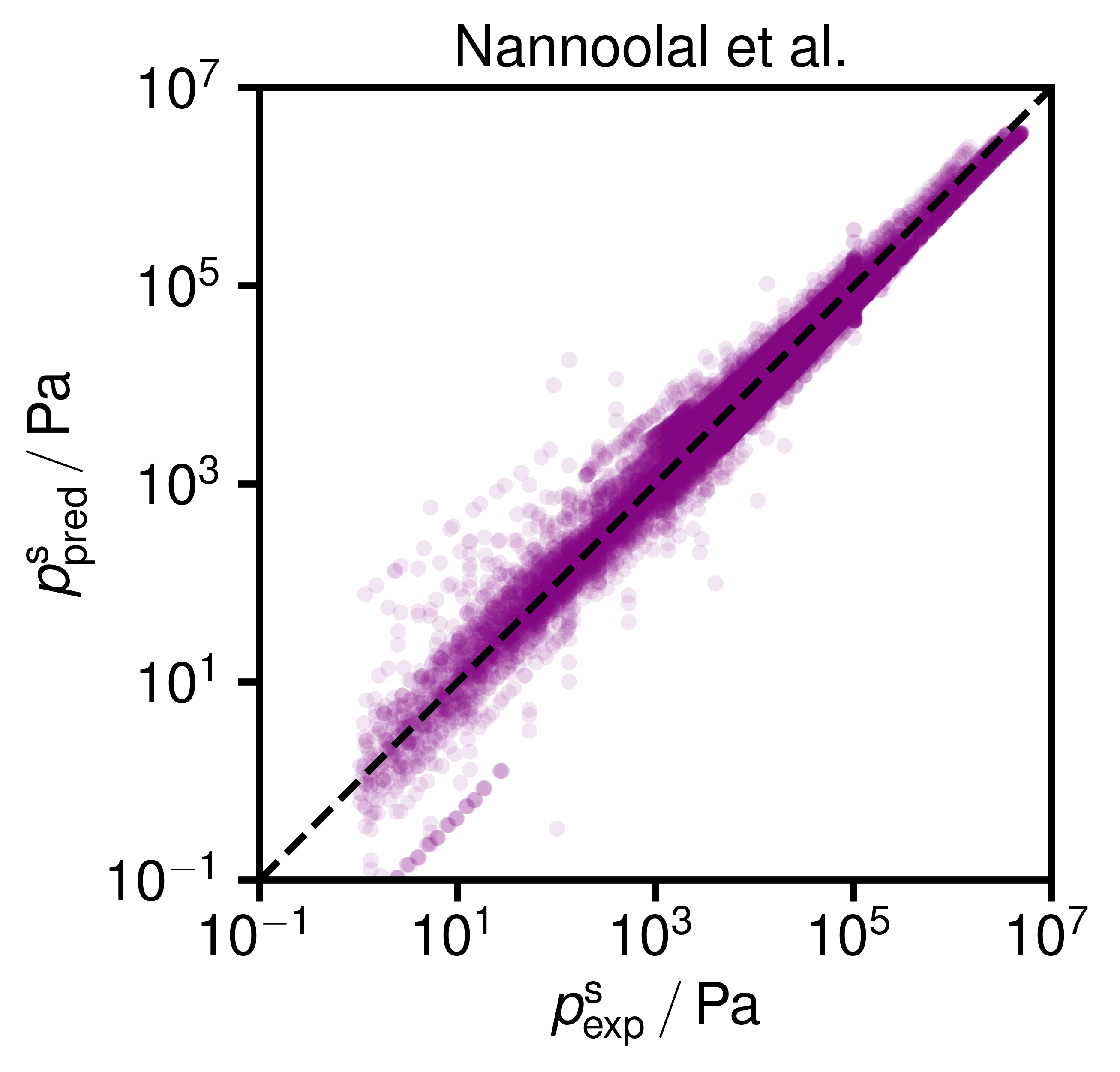}
    \end{subfigure}
    \hfill
    \begin{subfigure}{0.49\textwidth}
        \centering
        \includegraphics[width=\linewidth]{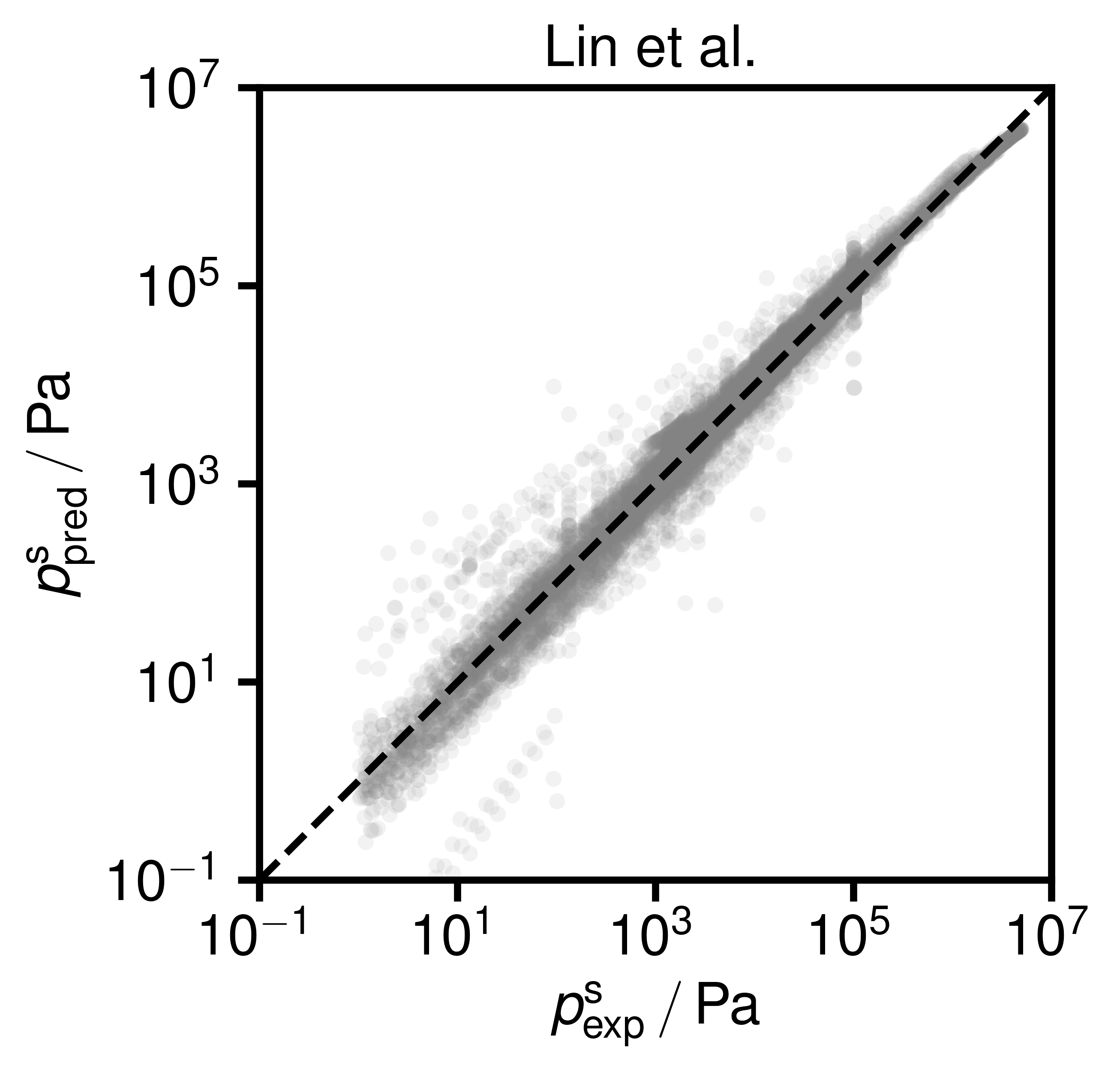}
    \end{subfigure}
    \hfill
    \caption{Parity plots showing the predicted vapor pressure for our test set over the experimental values for the method by Tu \cite{Tu1994}, SIMPOL \cite{Pankow2008}, the method by Nannoolal et al.\cite{Nannoolal2008}, and the method by Lin et al \cite{Lin2024}. Only results for components with at least two data points in the test set are shown. The dashed lines mark perfect predictions. Because of limitations of the literature models, predictions were only feasible for $36.1\,\%$~(Tu), $53.8\,\%$~(SIMPOL), and $94.9\,\%$~(Nannoolal et al.) of the components. The method by Lin et al. \cite{Lin2024} is applicable to our entire test set.}    \label{fig:evaluation_test_set}
\end{figure}

\begin{figure}
    \centering
    \includegraphics[width=0.8\linewidth]{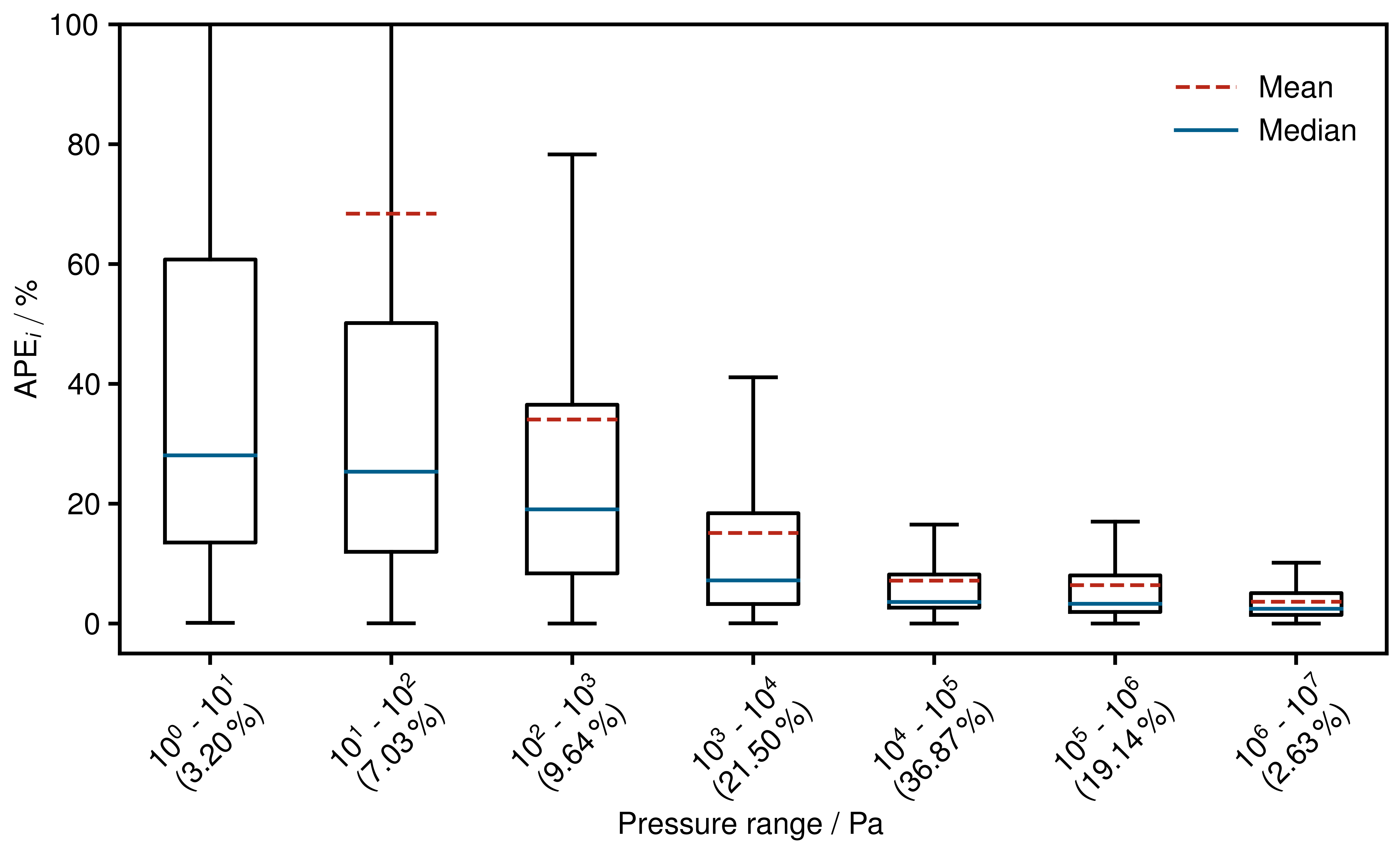}
    \caption{Boxplot visualizing the prediction accuracy of the developed GRAPPA model in terms of $\mathrm{APE}_i$ for different pressure intervals on the test set for components with at least two data points. The boxes represent the interquartile range, and the whiskers are 1.5 times the interquartile range. The numbers in the brackets denote the percentage of data points falling into the respective pressure range.}    \label{fig:boxplot_pressures}
\end{figure}

\begin{figure}
    \centering
    \includegraphics[width=0.8\linewidth]{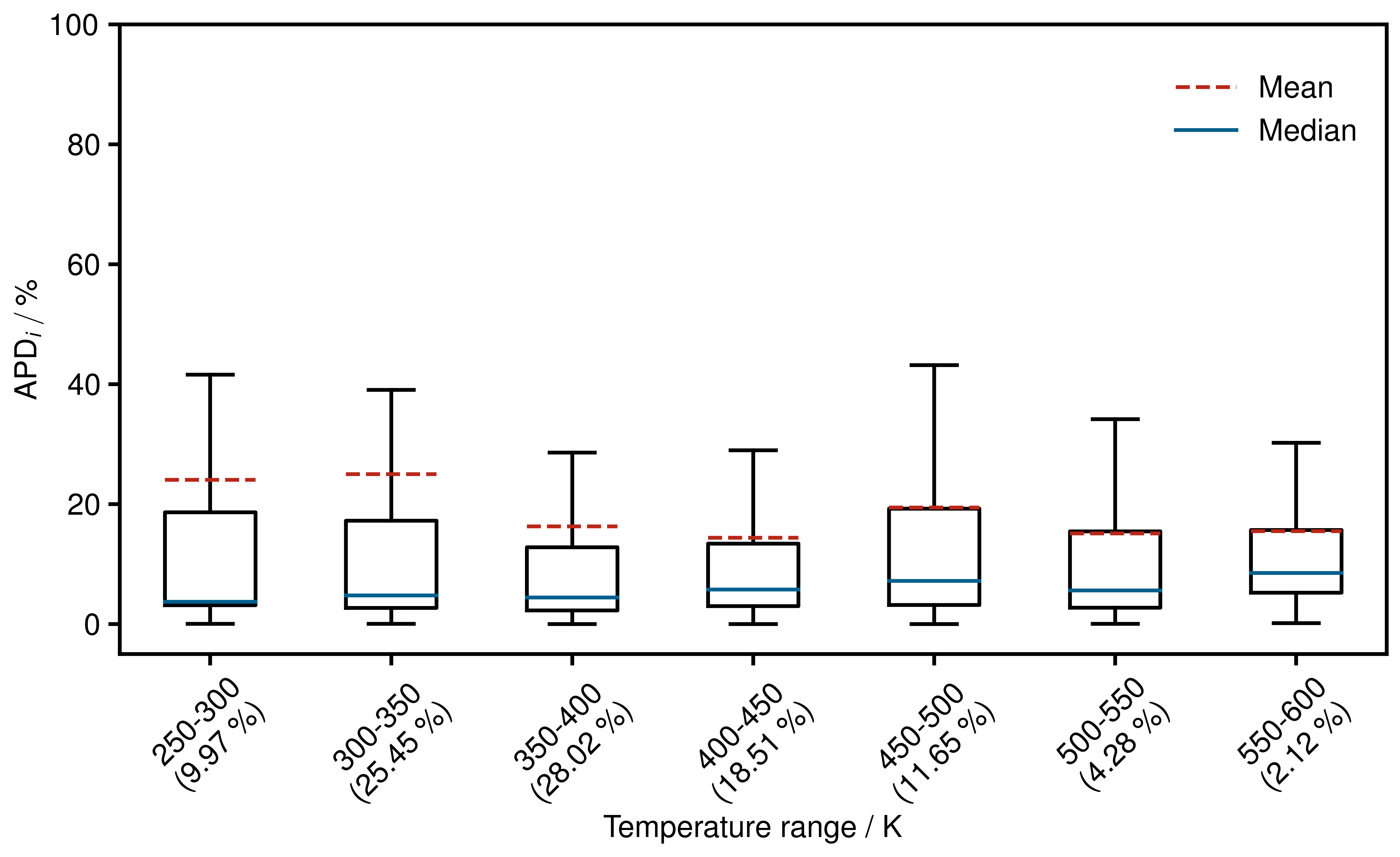}
    \caption{Boxplot visualizing the prediction accuracy of the developed GRAPPA model in terms of $\mathrm{APE}_i$ for different temperature intervals on the test set for components with at least two data points. The boxes represent the interquartile range, and the whiskers are 1.5 times the interquartile range. The numbers in the brackets denote the percentage of data points falling into the respective temperature range.}   
    \label{fig:boxplot_temperatures}
\end{figure}

\begin{figure}
    \centering
    \includegraphics[width=0.8\linewidth]{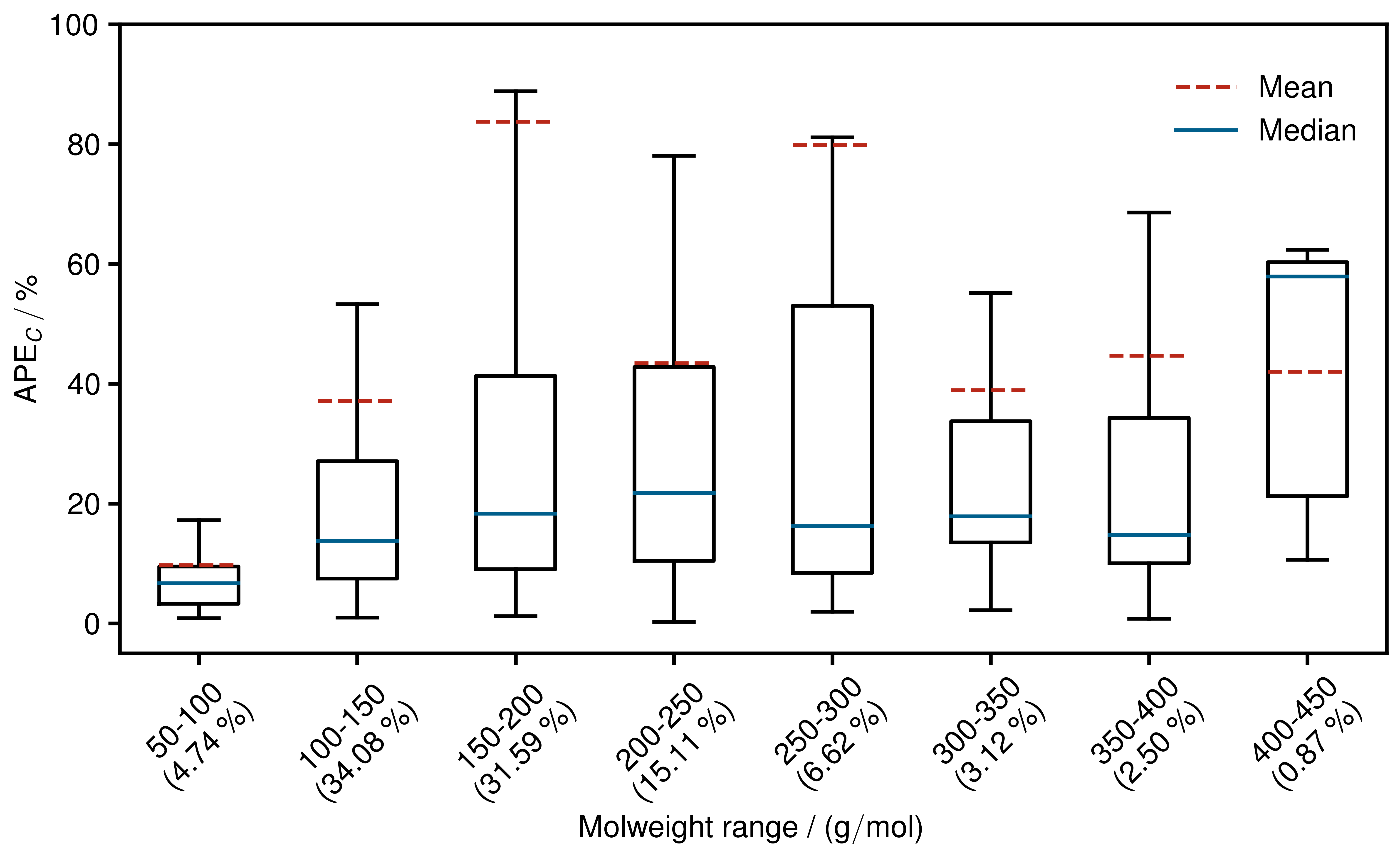}
    \caption{Boxplot visualizing the prediction accuracy of the developed GRAPPA model in terms of $\mathrm{APE}_C$ for different mol weight intervals on the test set for components with at least two data points. The boxes represent the interquartile range, and the whiskers are 1.5 times the interquartile range. The numbers in the brackets denote the percentage of data points falling into the respective mol weight range.}    \label{fig:boxplot_mol_weights}
\end{figure}

\begin{figure}
    \centering
    \includegraphics[width=0.8\linewidth]{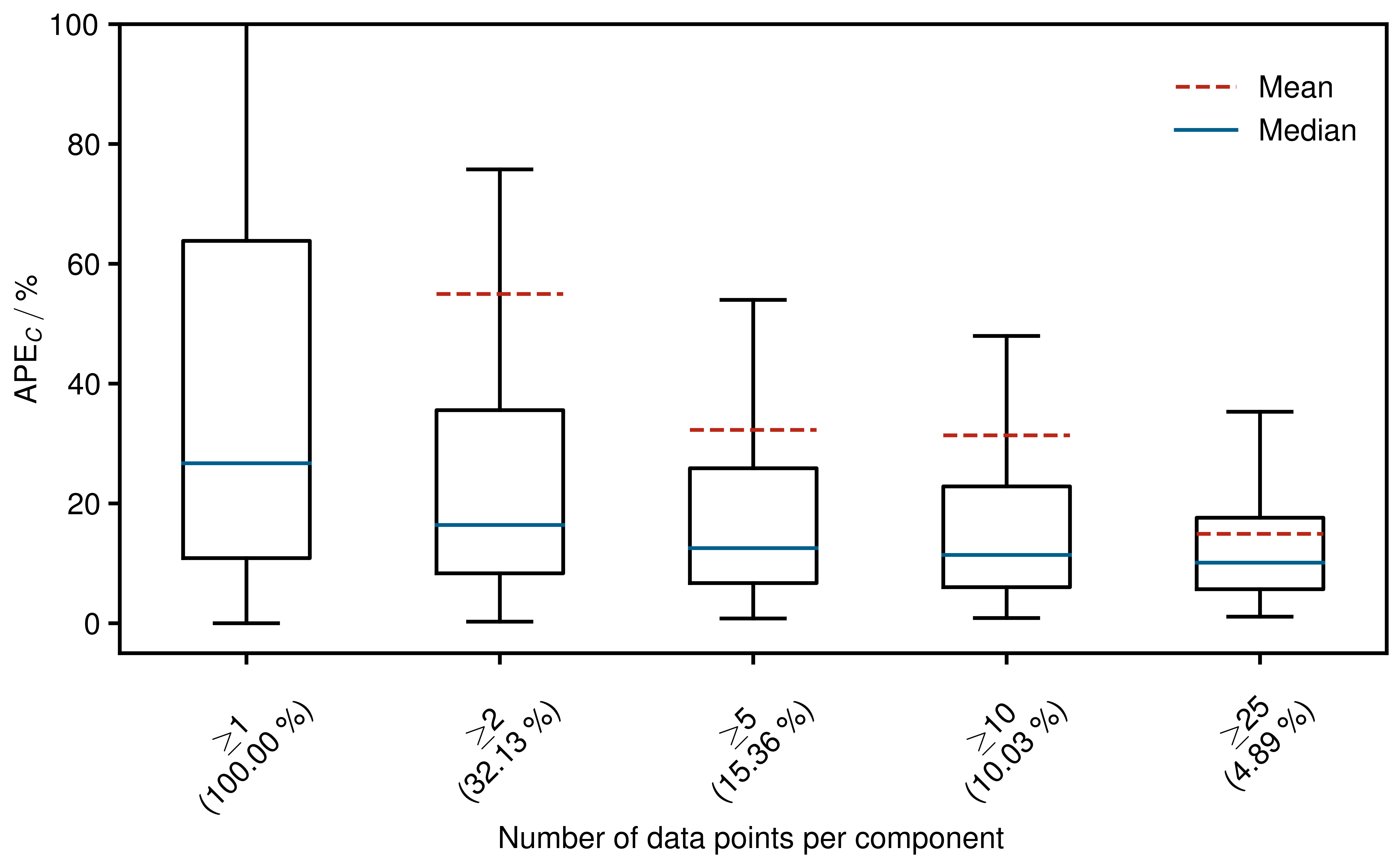}
    \caption{Boxplot visualizing the prediction accuracy of the developed GRAPPA model in terms of $\mathrm{APE}_C$ for different minimum numbers of data points per component in the test set. The boxes represent the interquartile range, and the whiskers are 1.5 times the interquartile range. The numbers in the brackets denote the percentage of components for which the minimum amount of data points are available.}    \label{fig:boxplot_data_points}
\end{figure}

The left panel of Fig.~\ref{fig:parity_plot_histogram_GRAPPA_Lin2024} shows a parity plot of the vapor pressure predictions of GRAPPA over the experimental data. In the right panel of Fig.~\ref{fig:parity_plot_histogram_GRAPPA_Lin2024}, the histogram shows the number of components predicted with a certain $\mathrm{APE}_C$ with GRAPPA and the results of the method by Lin et al.\cite{Lin2024} for comparison. In Fig.~\ref{fig:evaluation_test_set}, parity plots for the vapor pressure predictions of the four studied literature methods over the experimental data are shown. The pressure and temperature dependence of the prediction accuracy is further investigated in the boxplots in Fig.~\ref{fig:boxplot_pressures} and Fig.~\ref{fig:boxplot_temperatures}. The relationship between molecular weight and prediction accuracy is visualized in the boxplot in Fig.~\ref{fig:boxplot_mol_weights}. Fig.~\ref{fig:boxplot_data_points} shows how the prediction accuracy correlates with the number of experimental data points for a specific component in the test set.

\clearpage

\clearpage

\providecommand{\latin}[1]{#1}
\makeatletter
\providecommand{\doi}
  {\begingroup\let\do\@makeother\dospecials
  \catcode`\{=1 \catcode`\}=2 \doi@aux}
\providecommand{\doi@aux}[1]{\endgroup\texttt{#1}}
\makeatother
\providecommand*\mcitethebibliography{\thebibliography}
\csname @ifundefined\endcsname{endmcitethebibliography}  {\let\endmcitethebibliography\endthebibliography}{}